\documentclass[sigplan,screen]{acmart}

\usepackage{xspace}
\usepackage{mathtools}

\usepackage{bm}
\usepackage{enumitem}

\usepackage{textcomp}
\usepackage{xcolor}
\usepackage{changepage}
\usepackage{diagbox}

\usepackage{graphicx}
\usepackage{subcaption}
\usepackage{tikz}
\usepackage{caption}

\usepackage{tabularx}
\usepackage{multirow}
\usepackage{booktabs}
\usepackage{makecell}
\usepackage[figuresleft]{rotating}
\usepackage[flushleft]{threeparttable}

\usepackage{algorithmicx}
\usepackage{algorithm}
\usepackage[noend]{algpseudocode}

\algnewcommand\algorithmicinput{\textbf{Input:}}
\algnewcommand\Input{\item[\algorithmicinput]}
\algnewcommand\algorithmicoutput{\textbf{Output:}}
\algnewcommand\Output{\item[\algorithmicoutput]}

\usepackage[utf8]{inputenc} %
\usepackage{tcolorbox} %
\usepackage{soul} %
\usepackage{pifont} %
\usepackage{bbding} %
\usepackage[textsize=footnotesize, linecolor=magenta, bordercolor=magenta, backgroundcolor=white, textwidth=35pt]{todonotes}
\usepackage{marginnote}

\newcommand{\oney}{\raisebox{-0.35mm}{\Large{\ding{172}}}}
\newcommand{\twoy}{\raisebox{-0.35mm}{\Large{\ding{173}}}}
\newcommand{\threey}{\raisebox{-0.35mm}{\Large{\ding{174}}}}
\newcommand{\foury}{\raisebox{-0.35mm}{\Large{\ding{175}}}}
\newcommand{\fivey}{\raisebox{-0.35mm}{\Large{\ding{176}}}}
\newcommand{\sixy}{\raisebox{-0.35mm}{\Large{\ding{177}}}}

\usepackage[skip=6pt]{caption} 
\newcommand{\SysName}{\textrm{TLT}\xspace}

\definecolor{mydarkblue}{rgb}{0,0.08,1}
\definecolor{mydarkgreen}{rgb}{0.02,0.6,0.02}
\definecolor{mydarkred}{rgb}{0.8,0.02,0.02}
\definecolor{mydarkorange}{rgb}{0.40,0.2,0.02}
\definecolor{mydarkgold}{rgb}{0.6, 0.4, 0.05}
\definecolor{myred}{rgb}{1.0,0.0,0.0}
\definecolor{mygold}{rgb}{0.75,0.6,0.12}
\definecolor{myblue}{rgb}{0,0.2,0.8}
\definecolor{mylightblue}{rgb}{0.827,0.937,0.945}
\definecolor{mydarkgray}{rgb}{0.66,0.66,0.66}
\definecolor{mygray}{rgb}{0.85,0.85,0.85}

\copyrightyear{2026}
\acmYear{2026}
\setcopyright{cc}
\setcctype{by}
\acmConference[ASPLOS '26]{Proceedings of the 31st ACM International Conference on Architectural Support for Programming Languages and Operating Systems, Volume 2}{March 22--26, 2026}{Pittsburgh, PA, USA}
\acmBooktitle{Proceedings of the 31st ACM International Conference on Architectural Support for Programming Languages and Operating Systems, Volume 2 (ASPLOS '26), March 22--26, 2026, Pittsburgh, PA, USA}
\acmPrice{}
\acmDOI{10.1145/3779212.3790231}
\acmISBN{979-8-4007-2359-9/2026/03}

\begin{document}

\title{Taming the Long-Tail: Efficient Reasoning RL Training with Adaptive Drafter}

\settopmatter{authorsperrow=5}

\author{Qinghao Hu$^*$}
\affiliation{
  \institution{MIT}
  \city{Cambridge, MA}
  \country{USA}
}
\author{Shang Yang$^*$}
\affiliation{
  \institution{MIT}
  \city{Cambridge, MA}
  \country{USA}
}

\author{Junxian Guo}
\affiliation{
  \institution{MIT}
  \city{Cambridge, MA}
  \country{USA}
}

\author{Xiaozhe Yao}
\affiliation{
  \institution{ETH Zurich}
  \city{Zurich}
  \country{Switzerland}
}

\author{Yujun Lin}
\affiliation{
  \institution{NVIDIA}
  \city{Cambridge, MA}
  \country{USA}
}

\author{Yuxian Gu}
\affiliation{
  \institution{NVIDIA}
  \city{Cambridge, MA}
  \country{USA}
}

\author{Han Cai}
\affiliation{
  \institution{NVIDIA}
  \city{Cambridge, MA}
  \country{USA}
}

\author{Chuang Gan}
\affiliation{
  \institution{MIT-IBM AI Lab}
  \country{UMass Amherst}
}

\author{Ana Klimovic}
\affiliation{
  \institution{ETH Zurich}
  \city{Zurich}
  \country{Switzerland}
}

\author{Song Han}
\affiliation{
  \institution{MIT, NVIDIA}
  \city{Cambridge, MA}
  \country{USA}
}

\thanks{$*$\hspace{1mm}Equal Contribution. Part of the work was done while Shang Yang was interning at NVIDIA \vspace{-6pt}}

\renewcommand{\shortauthors}{Qinghao Hu et al.}



\begin{abstract}
    The emergence of Large Language Models (LLMs) with strong reasoning capabilities marks a significant milestone, unlocking new frontiers in complex problem-solving. However, training these reasoning models, typically using Reinforcement Learning (RL), encounters critical efficiency bottlenecks: response generation during RL training exhibits a persistent \emph{long-tail} distribution, where a few very long responses dominate execution time, wasting resources and inflating costs.
    To address this, we propose \SysName, a system that accelerates reasoning RL training losslessly by integrating adaptive speculative decoding. Applying speculative decoding in RL is challenging due to the dynamic workloads, evolving target model, and draft model training overhead. \SysName overcomes these obstacles with two synergistic components: (1) Adaptive Drafter, a lightweight draft model trained continuously on idle GPUs during long-tail generation to maintain alignment with the target model at no extra cost; and (2) Adaptive Rollout Engine, which maintains a memory-efficient pool of pre-captured CUDAGraphs and adaptively select suitable SD strategies for each input batch. Evaluations demonstrate that \SysName achieves over 1.7$\times$ end-to-end RL training speedup over state-of-the-art systems, preserves the model accuracy, and yields a high-quality draft model as a free byproduct suitable for efficient deployment. Code is released at \url{https://github.com/mit-han-lab/fastrl}.

\end{abstract}

\maketitle

\begin{figure}[!t]
    \centering
    \begin{subfigure}[b]{\linewidth}
        \includegraphics[width=\linewidth]{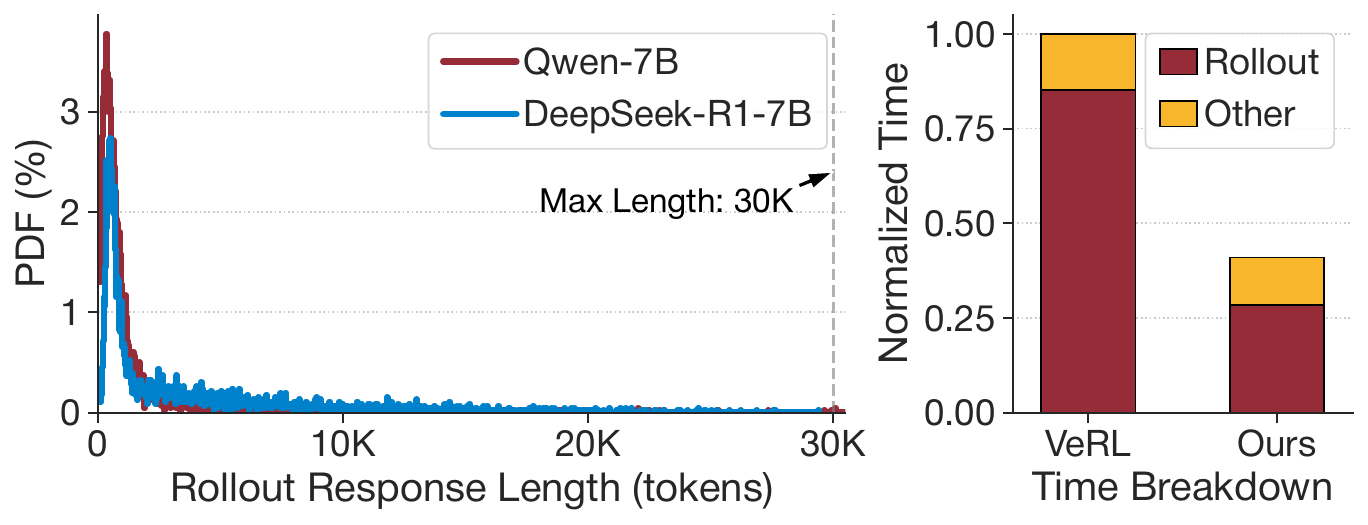}
        \caption{Distribution of response length and RL step time breakdown.%
        }
    \end{subfigure}
    \begin{subfigure}[b]{\linewidth}
        \includegraphics[width=\linewidth]{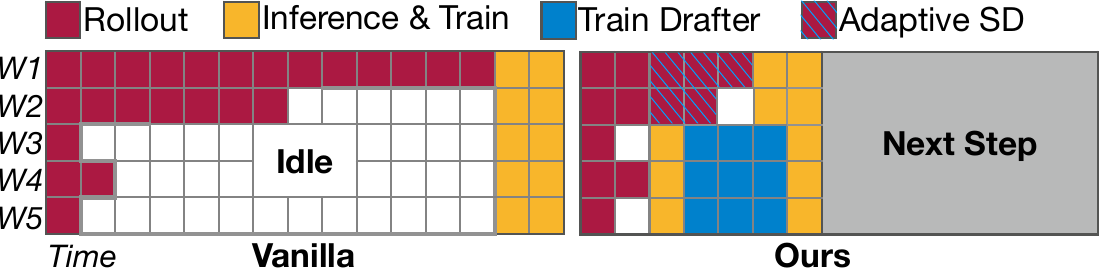}
        \caption{Illustration of the time line of RL reasoning step w./wo. \SysName.}
    \end{subfigure}
    \caption{Observed issues of long-tail generation (rollout) and workload imbalance in reasoning RL. \SysName system effectively addresses these challenges with adaptive drafter.}
    \label{figure_teaser}
\end{figure} 

\section{Introduction}
\label{sec_intro}

\begin{figure}[t]
    \centering
    \includegraphics[width=\linewidth]{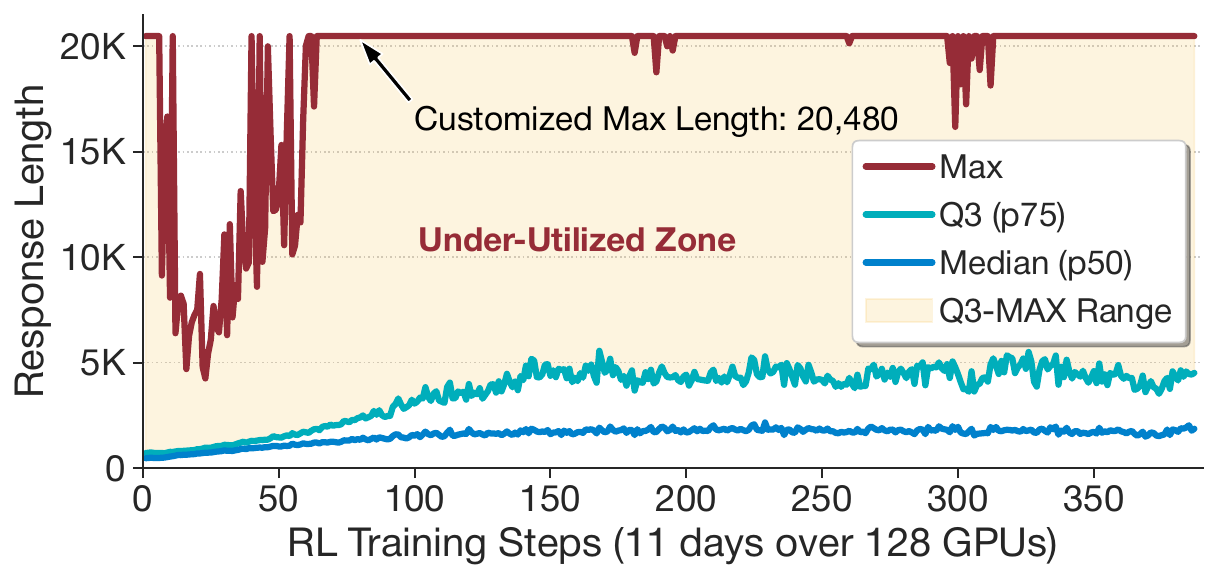}
    \caption{\textbf{RL Training trace from ByteDance} \cite{dapoTrace}, based on the Qwen2.5-32B model \cite{qwen25} and executed on H20 96GB GPUs. "p75" denotes the 75th percentile and more granular percentile data were not provided in the original source.}
    \label{figure_bytedance_trace}
\end{figure}

Over the past year, we have witnessed the emergence of reasoning LLMs, such as OpenAI-o1 \cite{O1}, DeepSeek-R1~\cite{DeepSeekR1} and Gemini-2.5Pro \cite{Gemini2.5Pro}, exhibiting remarkable capability improvements in mathematical reasoning, programming, and multidisciplinary knowledge tasks \cite{ReasoningSurvey}. A key driver behind these improvements is the use of reinforcement learning (RL), which has become the standard approach for endowing LLMs with strong reasoning capabilities~\cite{DeepSeekMath,QwQ,DeepSeekR1,DAPO}. However, this RL approach faces efficiency challenges due to the unique workload characteristics involved. Our analysis of self-collected traces (Figure~\ref{figure_teaser}) and  production traces from ByteDance (Figure~\ref{figure_bytedance_trace}) identifies the following key characteristics of reasoning RL process:

\noindent\textbf{(1) Unbalanced Rollout and Training Time}.
During RL, the first and most time-consuming stage is \emph{rollout}, where the model generates numerous candidate responses. These responses are then used in subsequent \emph{inference} and \emph{training} stages to update the model.
As shown in Figure \ref{figure_teaser} (a) , the rollout phase consumes a disproportionately large fraction ($\sim$85\%) of the total step time, becoming a primary bottleneck.

\noindent\textbf{(2) Persistent Long-Tail Distribution}.
A major source of RL training inefficiency arises from the \emph{long-tail distribution} of rollout response lengths. As shown in Figure~\ref{figure_teaser}(a), although most generated sequences are relatively short, a small fraction extends to extreme lengths. Crucially, this inefficiency is not an occasional occurrence but a persistent pattern observed throughout long-term training, as evidenced by ByteDance's production traces (Figure~\ref{figure_bytedance_trace}). In most RL steps, a few responses reach the maximum configured length, while the majority remain substantially shorter. The pronounced gap between this maximum length and the 75th percentile (p75) underscores significant resource under-utilization. 
Similar long-tail patterns have been independently reported in large-scale RL systems at Berkeley~\cite{deepscalerTrace}, Alibaba~\cite{gao2025rollpacker}, and AMD~\cite{zhou2025april}, indicating that this phenomenon is widespread.

\noindent\textbf{(3) Substantial Time and Resource Demands}. 
Figure~\ref{figure_bytedance_trace} shows that reasoning RL training for a 32B model took 11 days to complete only 385 steps, even using 128 GPUs. Individual training steps averaged approximately 40 minutes, while periodic evaluations (conducted every 5 steps) required roughly 20 minutes each, demonstrating the process is extremely time and resource-intensive. %

Existing systems \cite{HybridFlow, RLHFuse, ReaLHF, DeepSpeedChat, GEAR, NeMoAligner, FlexRLHF} for RLHF (RL from Human Feedback)  primarily focus their optimization efforts on managing multiple models, optimizing data transfer, and orchestrating device allocation across the different phases. However, they neglect tackling the rollout bottleneck, thereby remaining inefficient for reasoning RL tasks. Critically, reasoning rollouts are often over an order of magnitude longer than typical RLHF outputs \cite{HybridFlow,RLHFuse, InstructGPT}, giving reasoning RL fundamentally different workload characteristics and demanding distinct system-level optimizations.

While various acceleration techniques exist for LLM serving, many are ill-suited for the RL training context. Methods like quantization \cite{AWQ, GPTQ} and model sparsity \cite{LLM-Pruner, StreamingLLM}, despite offering speedups, are typically \emph{lossy}, risking model accuracy degradation and altering output probabilities. 
We identify Speculative Decoding (SD)~\cite{SpeculativeDecoding, chen2023accelerating} as a highly suitable approach. SD utilizes a lightweight draft model to generate token sequences speculatively, which are then verified in parallel by the main target model.
SD is particularly well-suited for RL training due to two primary merits: (1) \emph{Mathematically Lossless}. The output distribution remains identical to the target model's original distribution; (2) \emph{Efficient for Long-Tail}. SD enhances throughput by shifting the process from being memory-bound towards compute-bound, which is particularly effective for the long-tail phase of RL rollouts, where effective batch sizes are typically small.

However, existing speculative decoding techniques \cite{Medusa, Eagle1, SpecInfer, HASS, Falcon, REST} are designed for static inference scenarios, where the target model is fixed. Applying them within the dynamic reasoning RL training is fundamentally non-trivial, presenting significant challenges: (1) \emph{Evolving Target Model}: the target model's weights are constantly updated during RL training, making the draft model progressively stale, %
severely undermining the effectiveness of SD.
(2) \emph{Draft Model Training Costs}: Effective SD often relies on specialized draft models %
~\cite{HASS, Eagle1, Eagle2, Eagle3, Falcon, Hydra}, which require extra training to align with the target model, introducing additional overhead and complexity for users.
(3) \emph{Scheduling Complexity}: SD performance is sensitive to batch size and hyperparameters (e.g., draft depth). Reasoning RL rollouts, however, are characterized by dynamically fluctuating effective batch sizes as sequences complete at different times, necessitating an effective strategy selector for efficient SD deployment.

To address these challenges, we introduce \SysName, an efficient system for reasoning RL training featuring adaptive speculative decoding. \SysName aims to mitigate the long-tail rollout bottleneck while guaranteeing mathematical losslessness. 
The core design of \SysName derives from the following insights: 
First, \emph{Exploiting Rollout Bubbles}. The characteristically long duration of RL rollouts provides ample time to utilize GPU resources that progressively become available due to the long-tail nature. Resources freed as sequences complete can thus be repurposed for other tasks, such as draft model training, without additional costs \cite{Hydro}.
Second, \emph{Non-Blocking Drafter Training}. Draft model training can be decoupled from full rollout completion; training asynchronously on partial or readily available data allows its computation to be effectively overlapped with ongoing rollouts.

Incorporating aforementioned insights, we build \SysName with multiple system optimizations:
(1) Adaptive Drafter (\S \ref{sec_method_adaptive_drafter}):  We adopt a lightweight draft model continuously adapted via our spot trainer, innovatively addressing the evolving target model challenge arising from RL training. This trainer leverages idle GPU resources during long-tail rollouts for opportunistic, preemptible drafter updates, ensuring alignment with the evolving target model without adding overhead.
(2) Adaptive Rollout Engine (\S \ref{sec_method_rollout_engine}): \SysName maintains a memory-efficient pool of pre-captured CUDAGraphs and leverages the BEG-MAB tuner to adaptively select suitable SD strategies for each input batch. It also supports model-free speculative decoding for early rollout steps.
Figure \ref{figure_teaser}(b) presents an intuitive illustration of the \SysName workflow, where the system automatically leverages idle resources (e.g., workers W3-W5) for opportunistic draft model training (blue blocks), counteracting performance dips caused by target model updating without additional costs. Meanwhile, \SysName adaptively enables speculative decoding (blue hatched blocks) during these periods of resource under-utilization for better efficiency.

By synergistically combining an adaptive draft model with coordinated scheduling, \SysName effectively mitigates the long-tail issue inherent to RL training. We evaluated \SysName on various model architectures and real-world datasets. Experiments demonstrate that \SysName significantly accelerates end-to-end training throughput, while preserving model output quality and minimizing resource waste. Furthermore, the process yields an effective draft model suitable for future deployment as a free byproduct. 
\section{Background and Motivation}
\label{sec_motivation}

\subsection{Reasoning and Reinforcement Learning}
\label{subsec_background_reasoning}

\begin{figure}[t]
    \centering
    \includegraphics[width=\linewidth]{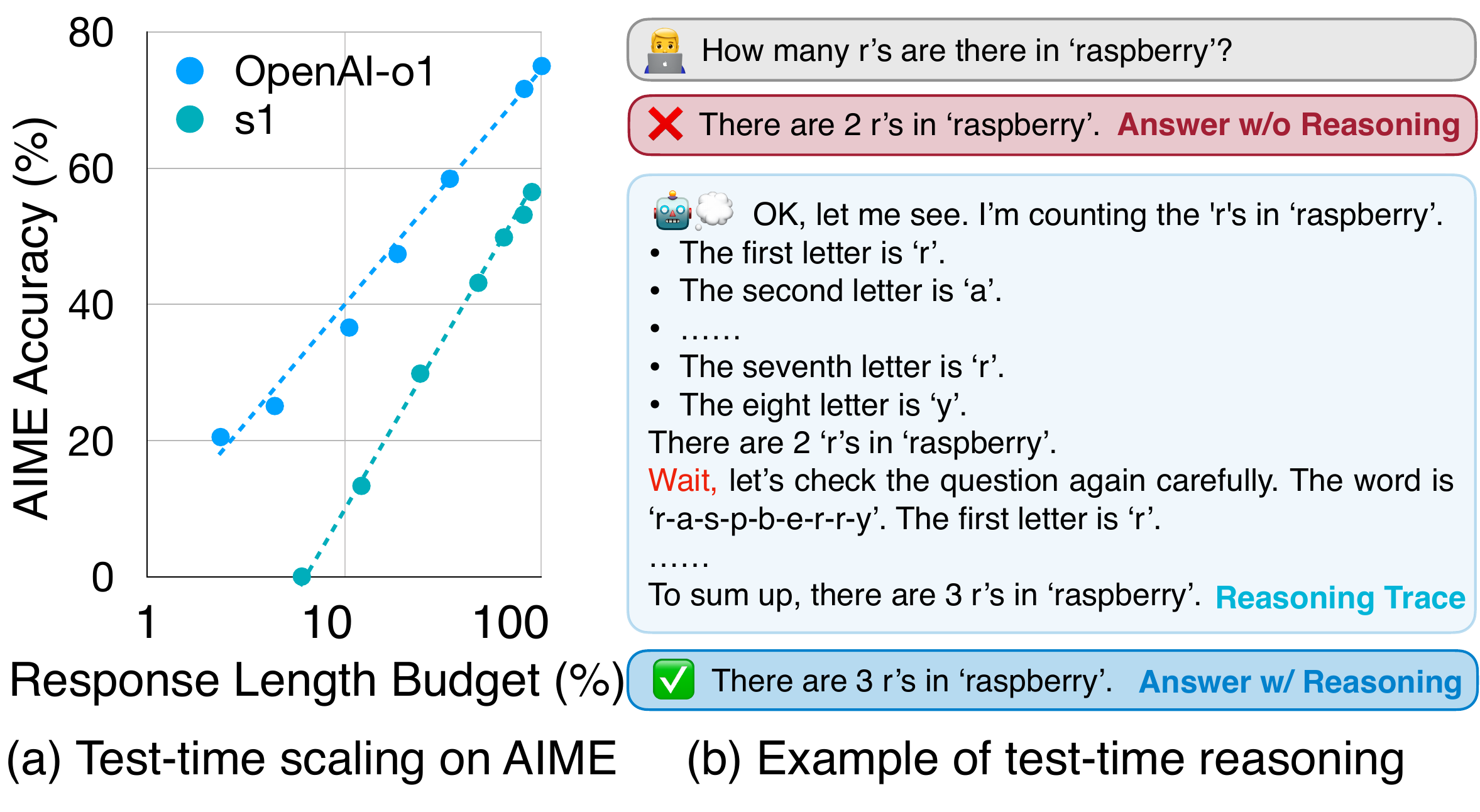}
    \caption{\textbf{Test-time scaling of reasoning models}. (a) Performance of OpenAI-o1~\cite{O1} and Stanford s1-32B~\cite{s1} on the AIME Competition-level Math Benchmark~\cite{AIME}. (b) Example of self-reflection correcting an error within the reasoning.
    }
    \label{fig:bg_and_motiv:test_time_scale}
\end{figure} %

\noindent\textbf{Reasoning LLMs}.
Large Language Models (LLMs) have recently shown remarkable performance in complex reasoning domains such as mathematics, logic, and programming. Powerful models such as OpenAI-o1~\cite{O1} and DeepSeek-R1~\cite{DeepSeekR1} achieve impressive accuracy on challenging reasoning benchmarks, including competition-level math and coding challenges. A key factor in their success is \emph{test-time scaling} \cite{InferenceScalingLaws, GreedyPolicySearch}, an inference paradigm that enhances accuracy by affording models more generation time to explore complex reasoning paths. 
These reasoning LLMs typically feature: (1) \emph{Long Chain-of-Thought}, generating extended intermediate reasoning steps before reaching a final answer; (2) \emph{Self-Reflection}, the capacity to evaluate, refine, and correct their own reasoning during inference to improve accuracy. 
As illustrated in Figure~\ref{fig:bg_and_motiv:test_time_scale} (a), allowing extended generation time enables these models to explore deeper and wider thinking traces~\cite{ReasoningSurvey}, thereby enhancing accuracy without additional training costs. More importantly, unlike traditional LLMs, these reasoning models can automatically revisit their generated responses and exhibit self-reflection capabilities (e.g., "\emph{wait}" in Figure~\ref{fig:bg_and_motiv:test_time_scale} (b) reasoning trace), which are typically acquired during their RL training stage \cite{DeepSeekR1} and are key to performance enhancements.


\begin{figure}[t]
    \centering
    \includegraphics[width=\linewidth]{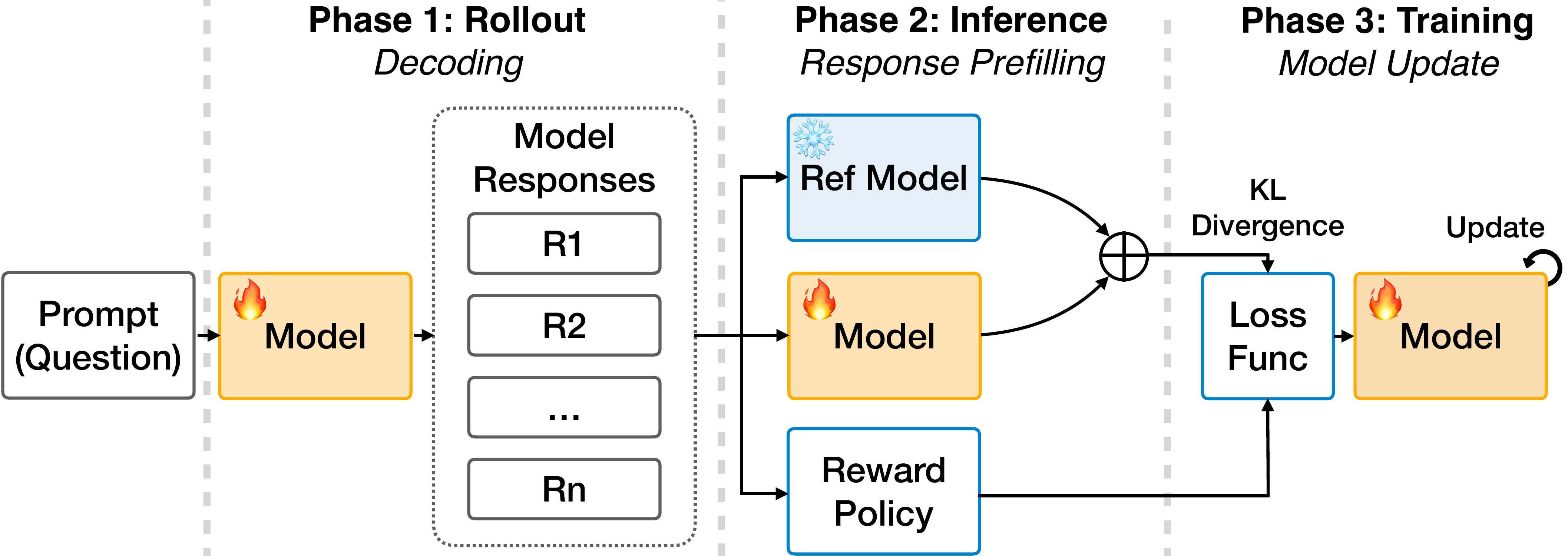}
    \caption{Overview of the GRPO \cite{DeepSeekMath} RL training process.}
    \label{fig:bg_and_motiv:rl_overview}
\end{figure} 
\noindent\textbf{Reinforcement Learning for Reasoning}. Recently, RL algorithm variants such as Group Relative Policy Optimization (GRPO)~\cite{DeepSeekMath} have proven their success in training reasoning LLMs. Unlike standard Policy Optimization (PPO) \cite{PPO}, which relies on an extensive critic model, GRPO employs a group-based baseline estimation to streamline the optimization process, significantly reducing training overhead while preserving the robustness of policy updates \cite{PostTrainingSurvey}.

Figure~\ref{fig:bg_and_motiv:rl_overview} illustrates a single training step in GRPO, which comprises three main stages: (1) \textbf{\emph{Rollout Stage}}, the target model generates multiple candidate responses for each input prompt; (2) \textbf{\emph{Inference Stage}}, the generated responses are processed by both the \textsl{target model} (the model undergoing RL training) and a fixed \textsl{reference model} (the initial target model, i.e., RL step=0). This processing yields logits used to compute a KL divergence penalty~\cite{john2020kld}, which serves to constrain model updates, ensuring stability and preventing excessive deviation from the reference model. Concurrently within this stage, a reward is calculated with a \emph{rule-based} policy regarding response characteristics (e.g., correctness), rather than relying on a separate value model; (3) \textbf{\emph{Training Stage}}, a loss function is constructed using the reward value and KL divergence. The target model's weights are then updated according to this loss.

\vspace{-10pt}
\subsection{Speculative Decoding for Efficient Reasoning RL}
\label{subsec_opportunities}

\begin{figure}[t]
    \centering
    \includegraphics[width=\linewidth]{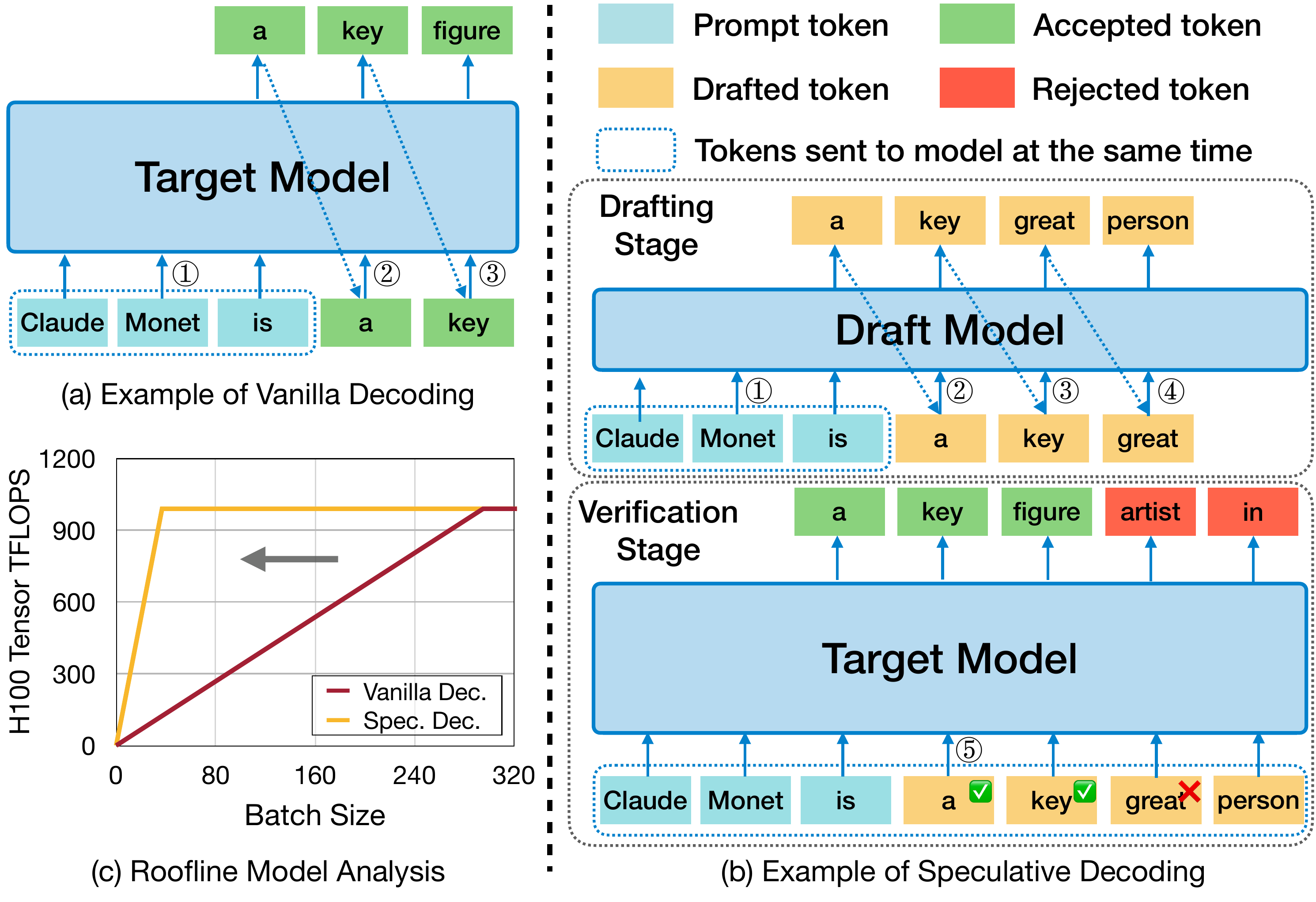}
    \caption{\textbf{Overview of Speculative Decoding.} (a \& b) Comparison between Vanilla and Speculative Decoding. (c) Speculative decoding achieves peak compute throughput (TFLOPS) at significantly smaller batch sizes (gray arrow). }
    \label{fig:bg_and_motiv:spec_dec}
    \vspace{-10pt}
\end{figure} 
As mentioned in \S\ref{sec_intro}, the long-tail rollout stage in reasoning RL frequently becomes a performance bottleneck. Consequently, tailored system optimizations for decoding acceleration are essential to improve throughput and resource efficiency. To this end, we identify speculative decoding~\cite{SpeculativeDecoding, chen2023accelerating} as a key approach well-suited to tackle these challenges.

\begin{figure*}[t]
    \centering
    \includegraphics[width=0.95\linewidth]{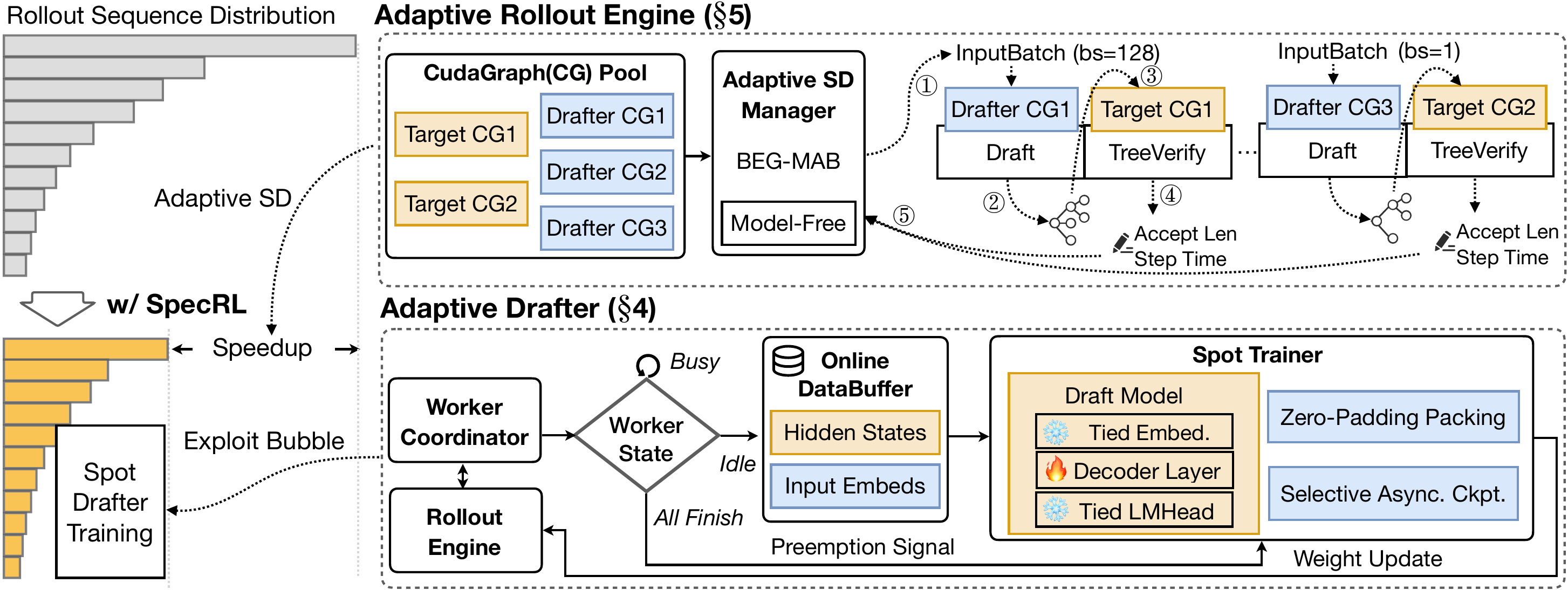}
    \caption{Overview of \SysName architecture and workflow.}
    \label{figure_system}
    \vspace{-10pt}
\end{figure*}

\noindent\textbf{Why Speculative Decoding}.
Among numerous methods accelerating LLM decoding, speculative decoding (SD) appears particularly advantageous for the reasoning RL scenario. Its primary benefit lies in its \emph{lossless} nature, mathematically ensuring the output distribution remains identical to that of the original target model. This fidelity is paramount in reasoning tasks where the correctness and logical coherence of generated steps are critical for effective reinforcement learning. 
It directly tackles the throughput bottleneck identified in the long-tail rollout stage by speeding up generation, enhancing sample efficiency without compromising the crucial accuracy needed for reliable reasoning RL training.

\noindent\textbf{Mechanism of Speculative Decoding}.
In speculative decoding (Figure~\ref{fig:bg_and_motiv:spec_dec} (b)), a lightweight draft model rapidly generates a sequence of candidate tokens. These candidates are then efficiently verified in parallel by the much larger target model in a single forward pass. The system accepts the drafted sequence up to the first token that mismatches the target model's prediction, also keeping the single correct token that the target model generated at that position. 
As shown in the roofline analysis in Figure~\ref{fig:bg_and_motiv:spec_dec} (c), this process improves throughput by shifting the typically memory-bound autoregressive generation towards a more compute-bound operation, more effectively saturating the GPU resources.

\noindent\textbf{Challenges in Reasoning RL}. 
Despite its effectiveness in alleviating the memory-bound issue and enhancing decoding throughput, adapting existing speculative decoding methods to the context of RL training is fundamentally non-trivial. We identify three key challenges:

\textbf{C1. Evolving Target Model}. Unlike standard inference scenarios where models are fixed, RL training involves continuous target model updates. This constant evolution renders the draft model used in SD progressively "stale", diminishing the acceptance rate of speculated tokens and significantly undermining SD's effectiveness.

\textbf{C2. Draft Model Training Costs}. 
Achieving high SD efficiency often requires dedicated draft models (e.g., single-decoder-layer models like Eagle \cite{Eagle1}, HASS \cite{HASS}), rather than arbitrary smaller models. Training a specialized drafter for the target model introduces substantial additional cost.

\textbf{C3. Fluctuating Batch Sizes}. 
As shown in Figure~\ref{fig:bg_and_motiv:spec_dec} (c), SD is typically optimized for small batch sizes. In contrast, RL rollouts often involve dynamically varying generation batch sizes. Applying SD efficiently in this scenario is challenging, as performance can degrade with larger batches if not managed properly, impacting overall throughput.

To address these challenges, \SysName leverages the unique characteristics of the reasoning RL training process. Specifically, we introduce the Adaptive Drafter (\S \ref{sec_method_adaptive_drafter}) to mitigate performance degradation from dynamic model weights (\textbf{C1}), utilizing spare GPU resources for efficient draft model updates and thereby removing training overhead (\textbf{C2}). Furthermore, we design a specialized scheduling strategy (\S \ref{sec_method_rollout_engine}) within the reasoning RL system to effectively manage the challenges posed by varying batch sizes (\textbf{C3}).

\vspace{-10pt}

\section{\SysName Overview}
\label{sec_sys_overview}

\noindent\textbf{Design Principles \& Goals}.
To facilitate practical adoption, \SysName adheres to four key design principles:

\noindent(a) \emph{Lossless Guarantee}. System optimizations must be lossless, preserving the mathematical equivalence in both rollout and training stages compared to the original algorithm, thereby avoiding asynchronous staleness or lossy approximations.

\noindent(b) \emph{Interference-Free}. The adaptive drafter workload must not interfere with the primary RL workload. Drafter updates are performed opportunistically (as spot tasks) and can be smoothly preempted to minimize impact on the RL workload.

\noindent(c) \emph{Automatic and Simple}. Manually setting up the draft model and configuring its speculative hyperparameters are burdensome and often lead to suboptimal performance. Thus, \SysName features an automated workflow for ease of use.

\noindent(d) \emph{Generalizable and Scalable}. The design must be adaptable to various popular reasoning RL algorithms (e.g., GRPO \cite{DeepSeekMath}, RLOO \cite{RLOO}). Furthermore, \SysName should scale effectively and deliver continuous system performance improvements across different model sizes, architectures, and cluster scales.

In addition, \SysName has three primary objectives: (1) Mitigate the long-tail problem to accelerate reasoning RL training; (2) Enhance cluster utilization and minimize resource waste; (3) Produce a high-performance draft model for future deployment with speculative decoding at no extra cost.

\noindent\textbf{Architecture \& Workflow}.
Figure~\ref{figure_system}  illustrates the overall architecture of \SysName and its workflow. The system is composed of two tightly integrated components: the Adaptive Drafter (\S \ref{sec_method_adaptive_drafter}) and the Adaptive Rollout Engine (\S \ref{sec_method_rollout_engine}).

The rollout process exhibits two key characteristics: a long-tail distribution of sequence lengths and progressively shrinking batch sizes as responses complete. To adapt to these dynamics, the Adaptive Rollout Engine maintains a pool of pre-captured CUDAGraphs for both target and draft models, managed by the Adaptive SD Manager. This manager leverages the proposed BEG-MAB tuner to automatically select appropriate SD strategies for each input batch. As shown in the top of Figure~\ref{figure_system}, the selected drafter CUDAGraph generates candidate tokens (\oney), which are then passed to the corresponding target CUDAGraph for parallel verification (\twoy-\threey). The engine measures acceptance length and step latency (\foury), feeding these signals back to the SD Manager (\fivey) to refine strategy selection in real time.

Meanwhile, the Adaptive Drafter module is responsible for continuously updating the lightweight draft model without interfering with the main rollout workload. A centralized Worker Coordinator monitors the state of rollout workers. When idle resources appear during the long-tail stage, the coordinator opportunistically launches Spot Trainer tasks. The training tasks use the Online DataBuffer, which caches hidden states and input embeddings from ongoing and prior steps' rollout. The draft model receives efficient updates using techniques such as zero-padding packing and selective asynchronous checkpointing. This design ensures that drafter updates can be preempted and resumed seamlessly, exploiting wasted bubbles.

By systematically combining adaptive rollout with opportunistic drafter training, \SysName achieves substantial speedups for reasoning RL without compromising model quality,  yielding a valuable draft model as a free by-product.

\vspace{-7pt}
\section{Adaptive Drafter}
\label{sec_method_adaptive_drafter}

\begin{figure}[t]
    \centering
    \includegraphics[width=\linewidth]{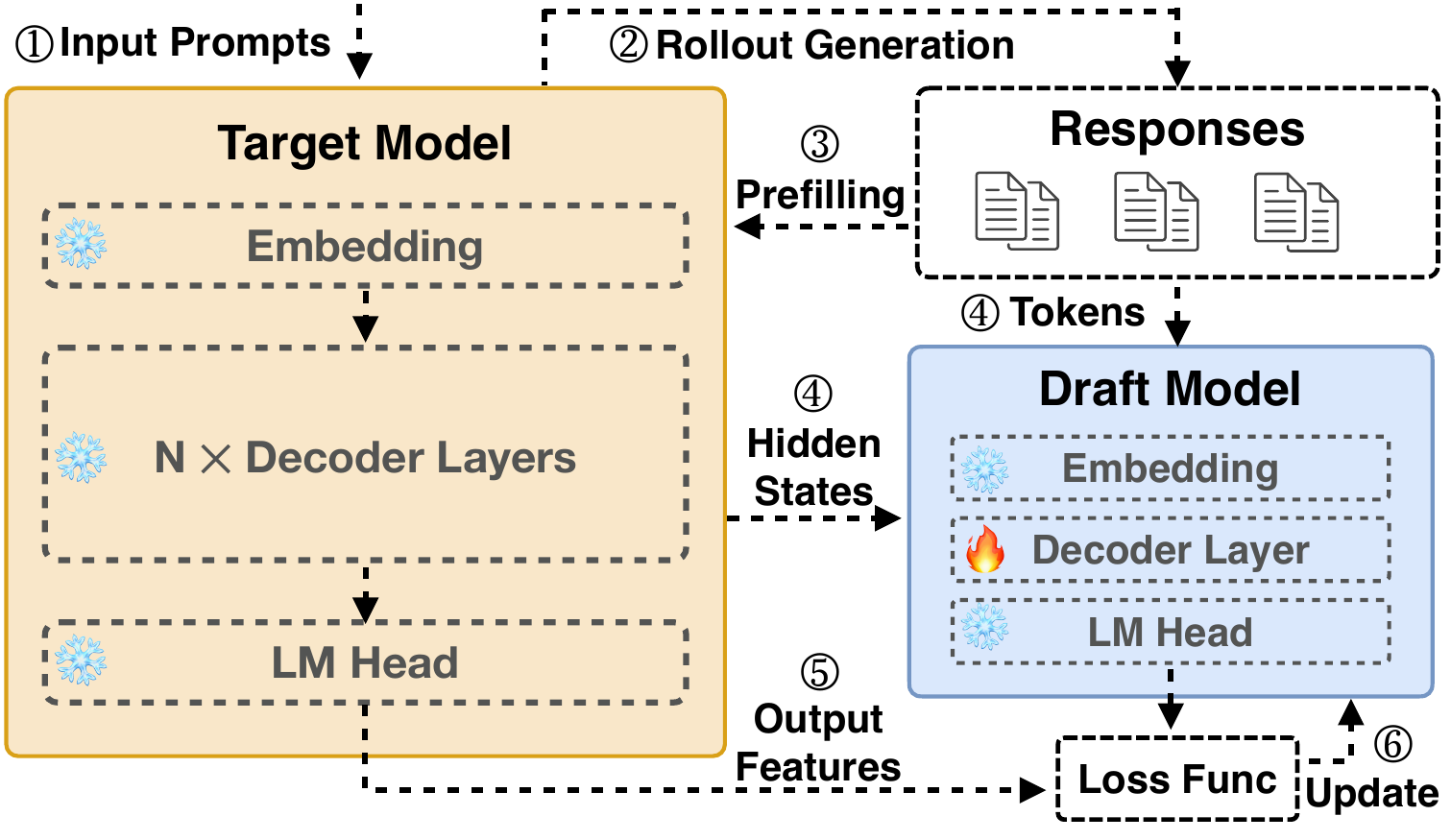}
    \begin{tabular}{@{}lccc@{}}

    \toprule
        Strategies    & \foury HiddenStates                                                          & \fivey LossFunc         & \begin{tabular}[c]{@{}l@{}} \sixy Training \\ Time Test\end{tabular} \\ \midrule
    Eagle \cite{Eagle1}   & Last layer                                                            & L1 + CE loss & N.A.                                                            \\
    HASS \cite{HASS}    & Last layer                                                            & L1 + CE loss & 3 steps                                                         \\
    Eagle-3 \cite{Eagle3} & \begin{tabular}[c]{@{}l@{}}Top, mid and\\  bottom layers\end{tabular} & CE loss only      & 7 steps                                                         \\ \bottomrule
    \end{tabular}
    \caption{\textbf{Draft Model Training}. \emph{Upper}: Unified training workflow using target model hidden states. \emph{Lower}: Specific configurations of different training strategies.}
    \vspace{-6pt}
    \label{fig:method:draft_trainer}
\end{figure}
 
This section introduces the Adaptive Drafter component of \SysName, focusing on how it creates, trains, and utilizes draft models to accelerate the reasoning RL process without interference with the primary RL workload.

\vspace{-5pt}
\subsection{Draft Model}
\label{subsec_learning_based_drafter}
\SysName utilizes a compact, specialized model trained to mimic the target LLM's token generation behavior, enabling efficient and accurate speculative execution.

\noindent\textbf{Model Architecture}.
To maximize efficiency, \SysName adopts a highly lightweight, \emph{single-layer} model design for the drafter \cite{Eagle1, HASS,Eagle3}. In this context, a "layer" refers to one complete transformer decoder block, including attention and feed-forward components. This approach differs from vanilla speculative decoding~\cite{SpeculativeDecoding, chen2023accelerating}, which often uses a separate, smaller multi-layer LLM from the same family (e.g., Qwen2.5-0.5B for a Qwen2.5-32B target). The vanilla approach has drawbacks: suitable and accurate small drafters may not exist for every target LLM, not to mention that these smaller models can still incur substantial drafting latency. For instance, while Qwen2.5-32B has 64 layers, Qwen2.5-0.5B still contains 24 layers. Although the parameter count is reduced significantly (64$\times$ in this case), latency remains dominated by the sequential computation across multiple layers. Experiments demonstrate that \SysName's single-layer draft model, sharing approximately the same parameter count as the Qwen2.5-0.5B drafter, is significantly faster (2.4$\times$).

As illustrated in Figure~\ref{fig:method:draft_trainer} (blue block), the \SysName drafter mirrors the target model's architecture but incorporates only a single, trainable decoder layer. It reuses the target model's original Embedding and LM Head layers, keeping their weights frozen during training. Consequently, only the single decoder layer's parameters are updated, representing a small fraction (approximately \texttt{1/layer\_num}) of the target model's total parameters. This architectural design significantly reduces both training and inference overhead. The effectiveness of such single-layer drafters is corroborated by prior work \cite{HASS, Eagle1, Eagle2, Eagle3, Falcon, Hydra}, which has shown their capability to closely align with the target model's auto-regressive distribution and achieve high acceptance rates for speculative tokens. During operation, the trained drafter takes hidden states from the target model and auto-regressively drafts subsequent tokens for verification.

\noindent\textbf{Training Process}.
\SysName builds a unified training framework that supports diverse draft model training strategies and integrates seamlessly with the RL workflow. As depicted in Figure~\ref{fig:method:draft_trainer}, the training process leverages data generated during the RL workflow. Specifically, input prompts are first processed by the target model to generate rollout responses (\oney, \twoy). During the subsequent prefilling (i.e., inference) phase, hidden states are collected from specific layers of the target model (\threey). These collected hidden states, concatenated with input token embeddings, are then passed through a lightweight linear layer for dimension reduction. Then the output features serve as input to the drafter's single decoder layer for training (\foury). Given the target model's significantly larger size (over 20$\times$ than the draft model), its prefilling cost can be far more expensive than one drafter training iteration. In \SysName, we cache the hidden states generated during the inherent RL inference phase to host memory and reuse them, thereby eliminating this overhead. 

\SysName also accommodates multiple training objectives. The framework can apply an L1 loss between the hidden states of the target and draft models to align their representations, a cross-entropy (CE) loss on the output logits to match token predictions, or a combination of both (\fivey). The computed loss is then backpropagated to update only the drafter’s decoder layer, while embeddings and the LM head remain tied to the target model (\sixy).

Additionally, several training strategies are supported within this unified pipeline in \SysName, as our framework is drafter-agnostic. For instance, EAGLE \cite{Eagle1} uses only the last layer's hidden states; HASS \cite{HASS} further enhance the training by feeding the output feature of draft model back into itself for multiple time (i.e., training-time test); EAGLE-3 \cite{Eagle3} further fuses multi-layer hidden states from the target model, aiming for better acceptance rates. Despite these differences, all these variants can be naturally expressed within our training framework. In this work, we use the EAGLE model as the default, as it offers high accepted lengths and faster convergence with much lower training cost.

\begin{figure}[t]
    \centering
    \includegraphics[width=\linewidth]{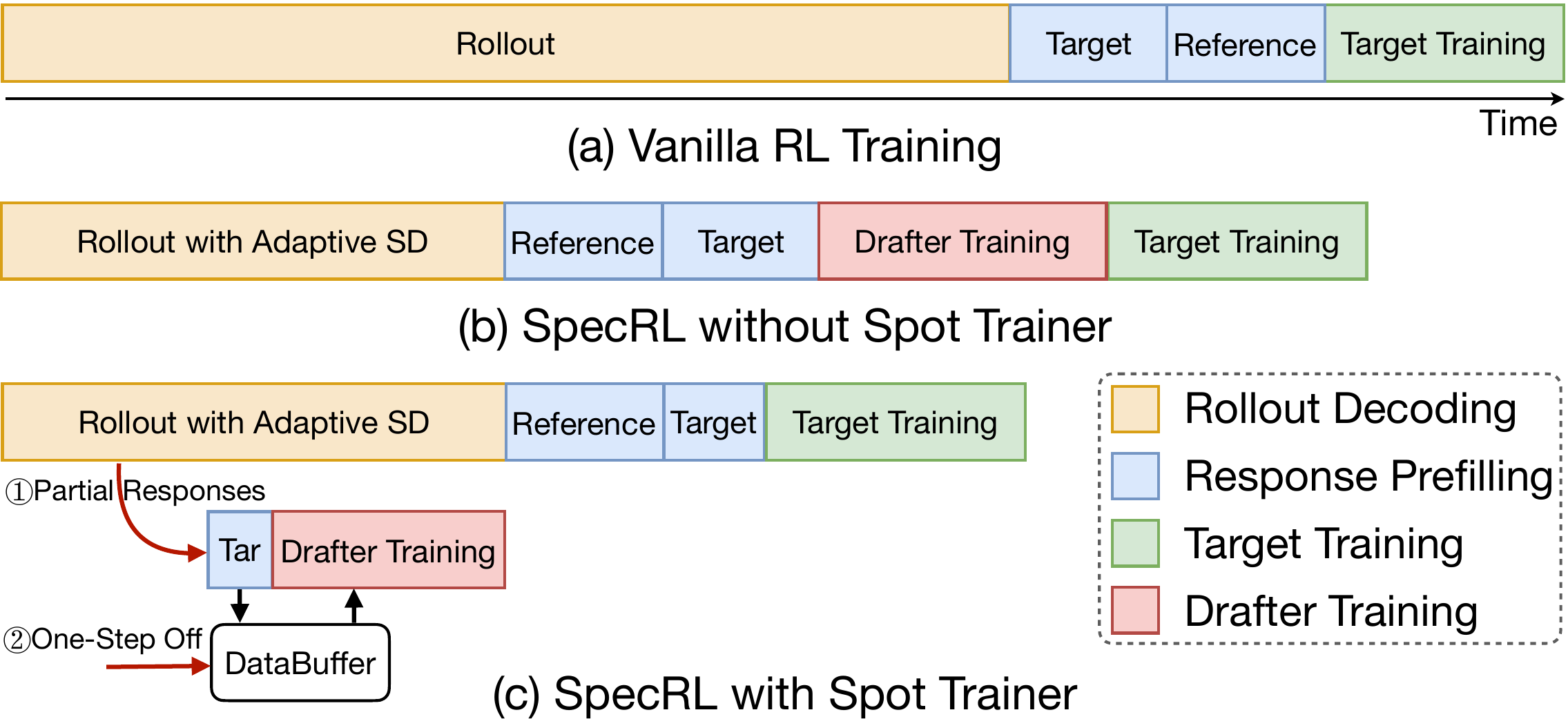}
    \caption{\textbf{Spot Trainer Workflow}. Block lengths are illustrative and not strictly proportional to execution time. Note that target model prefilling during drafter training is optional since hidden states from the rollout engine can be used directly.}
    \label{fig:method:waterfall}
\end{figure} 
\subsection{Spot Trainer}
\label{subsec_spot_updater}

The key challenge in applying speculative decoding within RL training is the \emph{evolving target model} (\textbf{C1}), causing a fixed draft model to quickly become stale and ineffective. Spot Trainer is designed to counteract this drift without hindering the primary RL workflow. To achieve this goal, \SysName incorporates the following novel designs:

\noindent\textbf{Worker Coordinator}.
It aims to bridge the gap between the Rollout Engine and the Spot Trainer, coordinating resource allocation, particularly repurposing GPU resources that become idle during the long-tail phase of rollout. In \SysName, a \emph{worker} is the basic unit of resource management, defined as one rollout instance. For example, when rollout uses a full DGX-H100 node with TP=8, a single worker represents 8 GPUs. The coordinator follows a centralized model implemented with ZeroMQ~\cite{ZeroMQ}, where a single coordinator process (rank 0) maintains the global state and workers communicate through asynchronous request-reply patterns. By monitoring active request counts and worker states, the coordinator determines when workers can be migrated from rollout to opportunistic drafter training. Each worker cycles between three states—\emph{BUSY} (serving rollout), \emph{IDLE} (inference finished and memory released), and \emph{TRAINING} (engaged in drafter updates)—and notifies the coordinator on every transition. Once the number of idle workers exceeds a configurable threshold, the coordinator promotes them to training mode and initiates drafter training.

Training initiation follows a leader-election pattern, where the first eligible worker sets up the training session and later workers can join if they belong to the same data-parallel group. When rollout completes, the coordinator immediately halts any ongoing drafter training and performs a graceful shutdown to ensure proper cleanup. This design ensures that idle GPU cycles during long-tail rollouts are effectively exploited without interfering with the primary RL workflow.

\noindent\textbf{Spot Training with DataBuffer}.
Figure~\ref{fig:method:waterfall} contrasts our Spot Trainer approach with standard RL workflows. Vanilla RL (a) involves a sequential process: rollout, target \& reference model inference for KL divergence calculation, and target model training (reward calculation is fast and omitted for brevity). A naive integration of drafter training (b) typically adds it as another sequential step, although reusing some prefilling computations but still incurs extra overhead.

In contrast, the Spot Trainer (c) operates on the principle that draft model training can proceed effectively without waiting for all generated responses to complete, allowing training to execute in a non-blocking manner. Due to the long-tail distribution in reasoning RL tasks, a majority of responses are generated significantly faster than the longest ones. The Spot Trainer leverages this observation by using partial of responses to train the draft model. However, relying only on early finishes means the drafter rarely sees very long sequences, which can introduce distribution mismatch. To address this, \SysName introduces the {DataBuffer}, which decouples drafter training from rollout completion. The DataBuffer caches hidden states and tokens produced during the inference stage and loads them with minimal overhead. Crucially, it supports \emph{one-step offset} sampling: we persist the buffer across RL steps and prioritize long sequences from the previous step to compensate for the scarcity of long-tail data in the current partial set. Although these samples are slightly stale, they remain effective for drafter training. Our experiments show that combining partial current-step data with buffered long sequences attains performance comparable to training on the full response set. Drafter updates are scheduled as low-priority “spot tasks” on idle GPUs, effectively harvesting compute freed by the long tail while keeping the main RL workflow non-blocking.

\begin{figure}[t]
    \centering
    \includegraphics[width=0.92\linewidth]{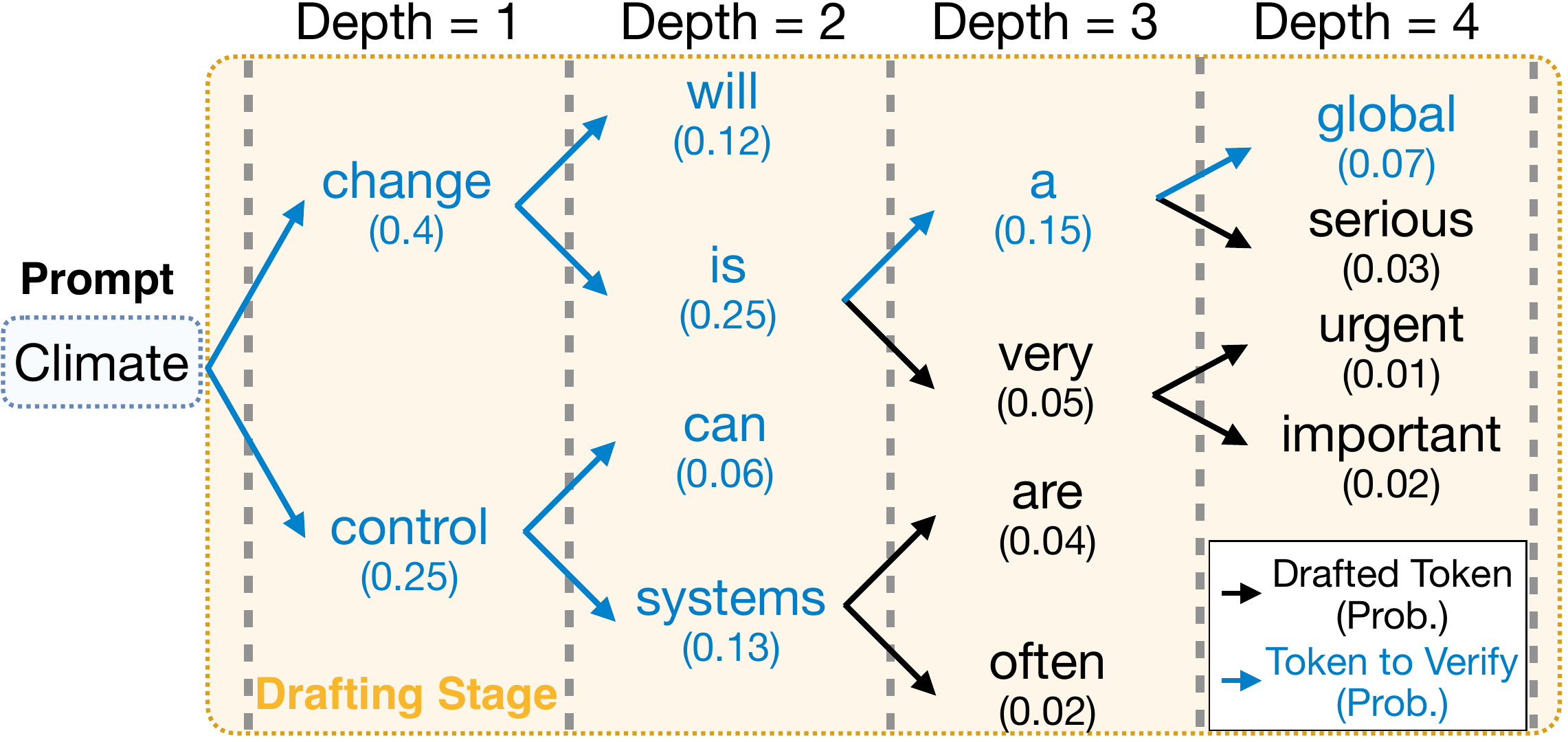}
    \caption{\textbf{Example of Tree Decoding}. We set \texttt{topK=2}, \texttt{Draft\_Depth=4}, and \texttt{Tokens\_to\_Verify=8} in this example. Prob. denotes token probability given by the draft model.}
    \label{figure_tree_decoding}
    \vspace{-10pt}
\end{figure} 

\noindent\textbf{Selective Asynchronous Checkpointing}.
A key design feature of the Spot Trainer is its preemptibility, allowing it to utilize idle resources without blocking the main RL workload. However, preemptions may lead to substantial loss of the draft model's training progress. To address this, the Spot Trainer utilizes asynchronous checkpointing: when triggered, the process of saving the draft model's state is offloaded to a background thread. Moreover, we optimize this process by filtering out frozen layers (e.g., embeddings and the LM head) and dumping only the trainable parameters, significantly reducing checkpointing latency. This mechanism minimizes work lost due to preemption through frequent and efficient checkpointing \cite{Acme}.

\noindent\textbf{Sequence Packing}.
The training data inherently consists of sequences with variable lengths. Standard batching techniques often require padding sequences to a uniform length, leading to inefficiency of computation, communication, and memory. To maximize GPU utilization during opportunistic drafter training, the Spot Trainer employs sequence packing \cite{SeqPacking}. This technique concatenates multiple variable-length sequences into a single packed sequence, removing padding while using attention masks to preserve sequence integrity during parallel processing. This enables efficient processing of the unbalanced training data within the limited, preemptible time slots available for spot training.

\vspace{-7pt}
\section{Adaptive Rollout Engine}
\label{sec_method_rollout_engine}

This section presents the Adaptive Rollout Engine, which dynamically adjusts SD strategies in response to real-time workload characteristics. It supports both learning-based and model-free SD, and incorporates a bucketed multi-armed bandit (MAB) tuner to automatically configure strategies.

\begin{figure}[t]
    \centering
    \includegraphics[width=\linewidth]{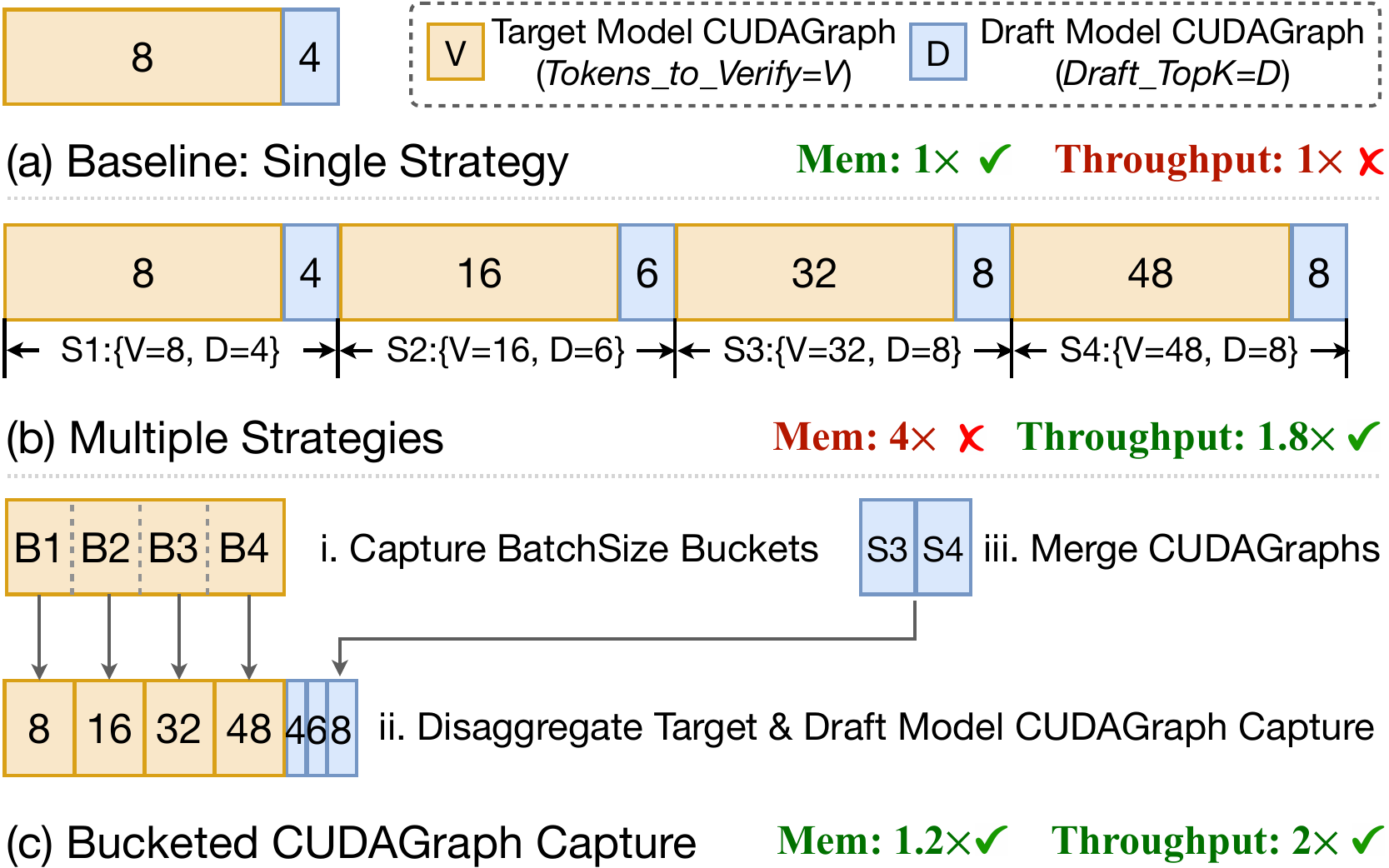}
    \caption{\textbf{Memory footprint of CUDA Graph for SD}. S1$\sim$S4 denote different SD strategies, B1$\sim$B4 represent batchsize buckets. Memory and throughput values are illustrative.}
    \label{fig:mab}
\end{figure}

\vspace{-7pt}
\subsection{Adaptive SD Manager}
\label{subsec_drafting_tuner}

\noindent\textbf{Tree-based Drafting}.
Unlike linear drafting, which processes a single sequence, tree-based drafting extends this single-branch paradigm by enabling parallel exploration of multiple speculative paths. 
Following~\cite{SpecInfer, Eagle2}, we build the candidate tree based on the draft model's confidence score (token probability), as illustrated in Figure~\ref{figure_tree_decoding}.
Starting from the input prompt ("Climate"), the draft model predicts potential next tokens. The tree-based approach explores the \texttt{topK} most likely options at each step (e.g., \texttt{topK}=2 yields "change" and "control"), and this branching continues for \texttt{Draft\_Depth} steps, creating a tree of candidate sequences. Finally, the \texttt{Tokens\_to\_Verify}=8 highest-confidence tokens within the tree are chosen and submitted to the target model for parallel verification. By exploring and verifying multiple potential sequences simultaneously, tree-based drafting substantially increases the number of accepted tokens per verification step, thereby improving overall generation throughput and GPU utilization. 

\noindent\textbf{Adaptivity is Necessary}.
While the SD is primarily designed to accelerate the long-tail stage, we observe that SD can also deliver speedups at moderate batch sizes (Table~\ref{tab:results:sd_bs}). However, this requires adaptive adjustment of SD strategies according to the decoding batch size and workload characteristics. Prior work~\cite{Eagle1,Eagle2,HASS} mainly focuses on batchsize=1 scenarios, which limits its applicability in real-world scenarios.
RL rollout naturally involve highly dynamic batch sizes: they begin large and then shrink rapidly as short responses complete. Static SD strategies are therefore both inefficient and unsafe, as they may cause Out-of-Memory (OOM) failures under large batches. To address this, the Adaptive SD Manager continuously tunes SD parameters—choosing safe strategies to prevent OOM, maximizing throughput by learning effective strategies for each workload, and balancing speculative gains against computational overhead. In addition, to handle large batch sizes where SD offers little benefit, it employs an elastic mechanism that activates SD only when the remaining request count drops below a configurable threshold (default: 32).

\noindent\textbf{Memory-Efficient CUDAGraph Capture}.
To accelerate LLM decoding, CUDAGraph is commonly adopted, providing substantial speedup by recording CUDA operations and replaying them as a single unit, which reduces per-kernel launch costs during inference. However, CUDAGraph requires capturing a separate static graph for each batch size and strategy, including both target and draft models. This inflexible paradigm leads to a considerable memory footprint. Under the Multiple-Strategy setting (Figure~\ref{fig:mab} (b)), memory usage grows linearly with the number of strategies, leaving insufficient space for model weights and KV caches, which is unacceptable in practice.

To accommodate more SD strategies without incurring prohibitive memory costs, we propose Bucketed CUDAGraph Capture (Figure~\ref{fig:mab} (c)), which incorporates three key optimizations: (1) Bucketed batch sizes. Instead of capturing a static graph for every possible batch size, we exploit strategy-specific characteristics (e.g., larger batch sizes typically require verifying fewer tokens) and group batch sizes into several buckets (e.g., B1$\sim$B4). 
(2) Disaggregated capture for target and draft models. Since some configurations affect only one model (e.g., \texttt{topK} impacts only the drafter, while \texttt{Tokens\_to\_Verify} affects only the target), we capture graphs for target and draft models separately to avoid multiplicative memory consumption. (3) Merged captures across strategies. When different strategies share identical parameter settings, we merge their batchsize buckets into a single captured graph, eliminating redundant memory overhead.
By integrating these optimizations, we substantially reduce the CUDAGraph memory footprint (Table~\ref{tab:results:cuda_graph_mem}) while improving rollout throughput. The throughput gain stems from leveraging prior knowledge to avoid evaluating suboptimal strategies that are unsuitable for specific batch sizes.

\begin{algorithm}[t]
    \caption{Bucketed-Epsilon-Greedy (BEG) MAB Selector}
    \small
    \label{alg:beg}
    \begin{algorithmic}[1]
        \Input \textbf{Strategies:} $\mathcal{S}$, \textbf{Batch Thresholds:} $\mathcal{T}=\{t_1,\dots,t_m\}$, \textbf{Exploration:} $\epsilon$, \textbf{Window Size:} $w$
        \Output Selected strategy $s^*$

        \Procedure{Initialize}{$\mathcal{S},\mathcal{T},\epsilon,w$}
            \State GroupByVerifyTokens($\mathcal{S}$) $\rightarrow$ $\{\mathcal{S}_1,\dots,\mathcal{S}_m\}$ (sorted by Tokens\_to\_Verify, descending)
            \State Define buckets $\mathcal{B}_i = [t_i, t_{i+1}\!-\!1]$ for $i<m$ and $\mathcal{B}_m=[t_m,\infty)$
            \State Map bucket $\mathcal{B}_i \mapsto$ group $\mathcal{S}_i$
            \ForAll{$s \in \mathcal{S}$}
                \State $R_s, A_s \gets$ deque ($w$) \Comment{init double-ended queue (size $w$)}
            \EndFor
        \EndProcedure

        \Procedure{Record}{~\texttt{elapsed\_time},~\texttt{accept\_lens},~$\texttt{batch\_size}$} 
            \State $\overline{a} \gets \big(\sum \texttt{accept\_lens}\big)/\texttt{batch\_size} + 1$
            \State $r_s \gets \overline{a} \times \texttt{batch\_size} / \texttt{elapsed\_time}$ \label{line:record}
            \State Append $r_s$ to $R_s$; Append $\overline{a}$ to $A_s$ \Comment{reward \& acc\_len}
        \EndProcedure

        \Procedure{SelectStrategy}{\texttt{batch\_size}}
            \State $V \gets \{\,s \in \mathcal{S}_i \mid \texttt{batch\_size}\in \mathcal{B}_i\,\}$ \Comment{candidate strategies}
            \If{$|V|=1$}
                \State \Return the unique $s\in V$
            \EndIf
            \State Draw $u \sim \mathrm{Uniform}(0,1)$
            \If{$u < \epsilon$}
                \State \Return random $s \in V$ \Comment{explore}
            \Else
                \State \Return $\arg\max_{s \in V} \mathrm{Median}(R_s)$ \Comment{exploit}
            \EndIf
        \EndProcedure
    \end{algorithmic}
\end{algorithm}

\subsection{Auto-Tune Algorithm}

\noindent\textbf{Bucketed $\epsilon$-Greedy MAB}.
Multi-Armed Bandit (MAB) algorithms are a classical framework for online decision-making, excelling at maximizing cumulative gains when selecting actions (“arms”) with uncertain rewards~\cite{MABResourceAllocation, MAB}. Each "arm" corresponds to a specific speculative decoding configuration tuple: {(\texttt{Draft\_Depth}, \texttt{topK}, \texttt{Tokens\_to\_Verify})}, and the "reward" reflects the efficiency within the generation step (i.e., $\text{accepted\_token\_num} / \text{step\_latency}$). 

To automate the selection of SD strategies, we design a new online tuning algorithm, Bucketed-Epsilon-Greedy (BEG) MAB Selector (Algorithm~\ref{alg:beg}). BEG adopts an $\epsilon$-greedy policy tailored to speculative decoding workloads. It buckets strategies according to their \texttt{Tokens\_to\_Verify} and dynamically match them to batchsize ranges. Specifically, strategies $\mathcal{S}$ are grouped by \texttt{Tokens\_to\_Verify} into ${\mathcal{S}_1,\dots,\mathcal{S}_m}$, and each group is mapped to a batch-size bucket $\mathcal{B}_i$. During rollout, the current batch size determines the relevant candidate set $V$. If $|V|=1$, the strategy is fixed; otherwise, BEG selects between exploration and exploitation. With probability $\epsilon$, a strategy is drawn uniformly at random from $V$ (exploration). With probability $1-\epsilon$, BEG selects the strategy maximizing the median reward over a sliding window of size $w$ (exploitation). The reward (line: \ref{line:record}) balances accept rate and latency cost. Maintaining deques of recent rewards ensures adaptation to non-stationary dynamics across RL training.

By combining bucketed strategy selection, lightweight $\epsilon$-greedy exploration, and sliding-window reward estimation, BEG reduces the overhead of exhaustive exploration while adapting efficiently to workload variations. This enables \SysName to scale across diverse batchsizes and speculative decoding strategies without manual tuning.

\vspace{-10pt}
\subsection{Model-free Drafter}
\label{subsec_model_free_drafter}

\noindent\textbf{Leveraging Similarity across Rollouts}. During the RL rollout stage, candidate responses generated for the same prompt often exhibit evident \emph{sequence similarity} and contain repetitive patterns, such as common mathematical notation or code syntax structures. This observation suggests that frequent, short-range token dependencies can be predicted effectively using retrieval methods. To leverage this, \SysName incorporates a complementary, non-parametric drafting strategy alongside the primary learning-based drafter. This model-free drafter operates by dynamically building an n-gram retrieval database populated from the rollout responses specific to each prompt. Based on the immediate context (the preceding n-grams), it efficiently predicts common subsequent token sequences by querying this database.

Within \SysName, the model-free drafter serves as both a robust baseline and an essential fallback mechanism. The Strategy Selector dynamically activates this retrieval-based drafter whenever the learning-based drafter is unavailable (e.g., during initial training steps) or predicted to be inefficient. This dual-drafter approach ensures that speculative decoding acceleration remains active throughout all stages of the RL process, contributing to persistent efficiency gains.

\vspace{-10pt}
\section{Evaluation}
\label{sec_evaluation}

\begin{figure*}[t]
    \centering
    \vspace{-10pt}
    \includegraphics[width=1.0\textwidth]{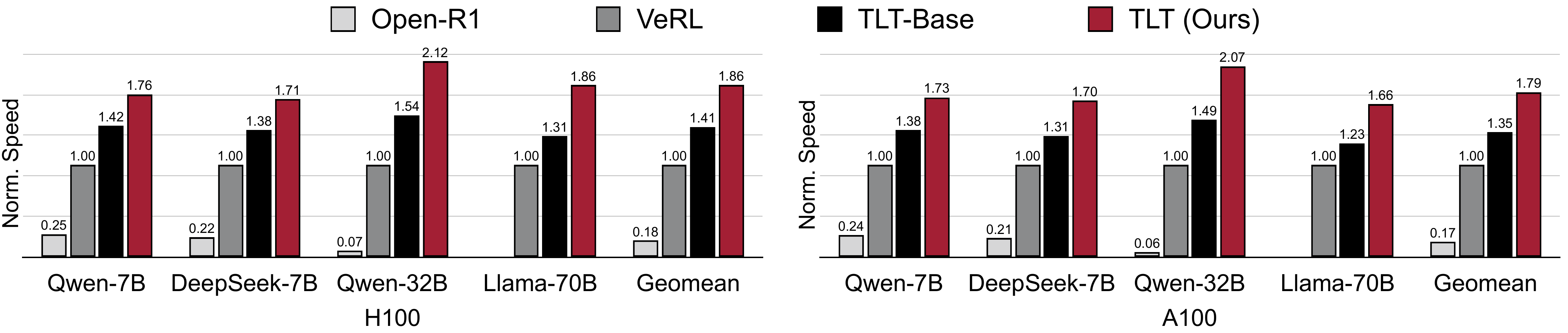}
    \caption{\textbf{End-to-end Training Speed Evaluation.} The y-axis indicates the relative training throughput of each system running GRPO~\cite{DeepSeekMath} RL algorithm. \SysName achieves 1.7-2.1$\times$ speedup over the state-of-the-art RL training system VeRL~\cite{HybridFlow}.
    }
    \label{fig:results:main}
\end{figure*}

\begin{figure*}[t]
    \centering
    \vspace{-10pt}
    \includegraphics[width=1.0\linewidth]{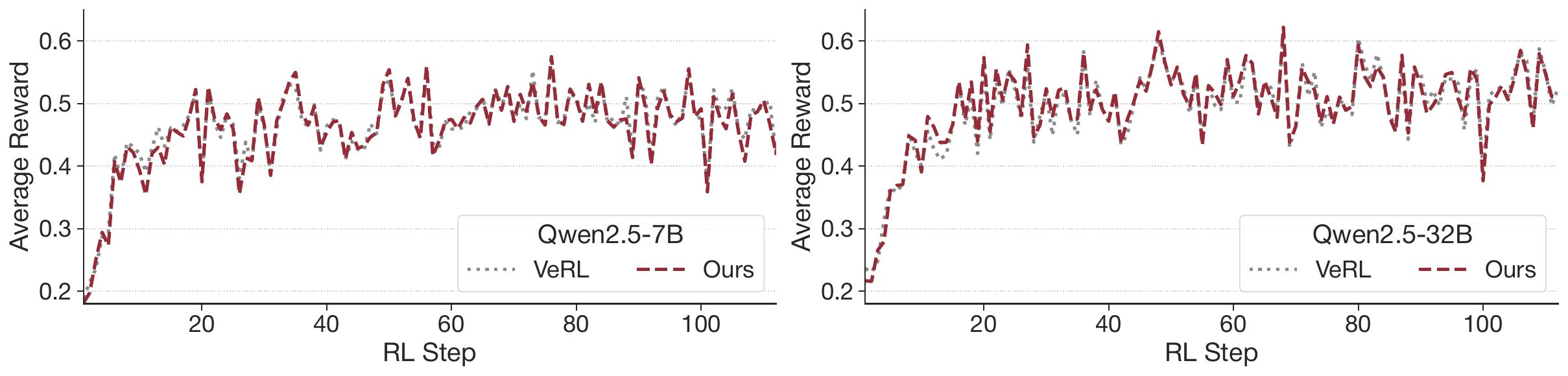}
    \caption{\textbf{End-to-end Training Curves.} Average reward curves for VeRL and \SysName using the Qwen2.5 7B and 32B models.  }
    \label{fig:results:rl_trace}
    \vspace{-10pt}
\end{figure*}

\subsection{Experimental Setup}
\label{subsec_setup}

\noindent\textbf{Implementation}.
We implement \SysName on top of VeRL \cite{HybridFlow} and utilizes Ray \cite{Ray} for distributed execution. The Worker Coordinator is realized with a centralized controller that employs ZeroMQ~\cite{ZeroMQ} messaging. The Adaptive Drafter's implementation is based on training pipelines from EAGLE and trained with FSDP2 \cite{FSDP2}. For asynchronous checkpointing, we leverage PyTorch DCP~\cite{DCP}. 
To further improve generation efficiency, we employ SGLang \cite{SGLang} as the rollout backend, with implementing adaptive enabling of SD, drafter weight update and the BEG-MAB tuner following~\cite{SGLangMAB}. We additionally use \cite{LookaheadKDD} as our model-free drafter.

\noindent\textbf{Testbed}.
We conduct experiments on a cluster of 8 NVIDIA DGX H100 \cite{dgx-h100}  servers, totaling 64 GPUs. Each server is equipped with 8 NVIDIA H100 GPUs, 2 Intel Xeon Platinum 8480C (112 cores) CPUs and 2TB memory. GPUs are interconnected to each other by 900GB/s NVLink, and inter-node communication is achieved via NVIDIA Mellanox 400Gb/s InfiniBand. To evaluate system's performance across heterogeneous hardware, we also conduct experiments on a different cluster equipped with NVIDIA DGX A100 servers. Unless otherwise specified, experiments are conducted on H100 GPUs by default.

\noindent\textbf{Models}.
Following the training procedure introduced in DeepSeek-R1~\cite{DeepSeekR1}, the reasoning RL process can begin either from a fine-tuned model or directly from a base model without additional fine-tuning. To demonstrate that \SysName robustly accommodates different model architectures, scales, and training paradigms, we evaluate the following models: 

\begin{enumerate}[label=(\arabic*),topsep=0pt,itemsep=0pt,parsep=0pt,leftmargin=20pt]
    \item \emph{Qwen2.5-7B} (Qwen2.5-7B): a popular base model;
    \item \emph{DeepSeek-R1-Distill-Qwen-7B} (DeepSeek-7B): a distilled model exhibiting good initial reasoning capabilities; 
    \item \emph{Qwen2.5-32B} (Qwen-32B): a larger base model without any additional fine-tuning, as in~\cite{DAPO}; 
    \item \emph{Llama-3.3-70B-Instruct} (Llama-70B): a larger instruction-tuned model to assess scalability.
\end{enumerate}

\noindent\textbf{Datasets}.
For the primary RL training dataset, we use a subset from Eurus-2-RL~\cite{Prime}, a comprehensive dataset curated for reasoning RL. This dataset contains mathematics and coding problems equipped with verifiers (LaTeX answers and test cases, respectively). The coding problems originate from competitive programming sources (e.g., CodeContests~\cite{CodeContests}, Codeforces~\cite{codeforces}), while the math problems range from high school to olympiad-level difficulty~\cite{numina_math}. For drafter training, we use a subset from OpenThoughts2-1M \cite{OpenThoughts2-1M} for warm-up.

\noindent\textbf{RL Settings}.
We primarily follow the reasoning RL algorithm GRPO proposed by DeepSeek \cite{DeepSeekR1, DeepSeekMath}. Specifically, for each prompt, we set the temperature=0.9, with a maximum generation length of 32K tokens, consistent with common RL training settings. We apply TP= 2, 4, or 8 depending on model scale for the rollout engine. Additionally, all experiments utilize mixed-precision training with the Adam optimizer \cite{Adam}, where BF16 precision is employed.

\noindent\textbf{Baselines}.
We consider the following baselines.
\begin{enumerate}[label=(\arabic*),topsep=0pt,itemsep=0pt,parsep=0pt,leftmargin=20pt]
    \item \emph{Open-R1} \cite{openr1}: a popular RL training framework built upon TRL\cite{TRL}, integrating vLLM~\cite{vLLM} for rollouts and DeepSpeed~\cite{ZeRO} for training. Currently, it only supports separate model placement, requiring serving and training to be performed on distinct nodes.
    \item \emph{VeRL} \cite{HybridFlow}: the state-of-the-art open-source RL training framework by ByteDance. It enables colocation of models on shared devices by utilizing GPU time-sharing.
    \item \emph{\SysName-Base}: Our base implementation with vanilla speculative decoding support. Specifically, we disable adaptive drafter and use the model-free drafter instead.
\end{enumerate}

\begin{figure}[t]
    \centering
    \includegraphics[width=\linewidth]{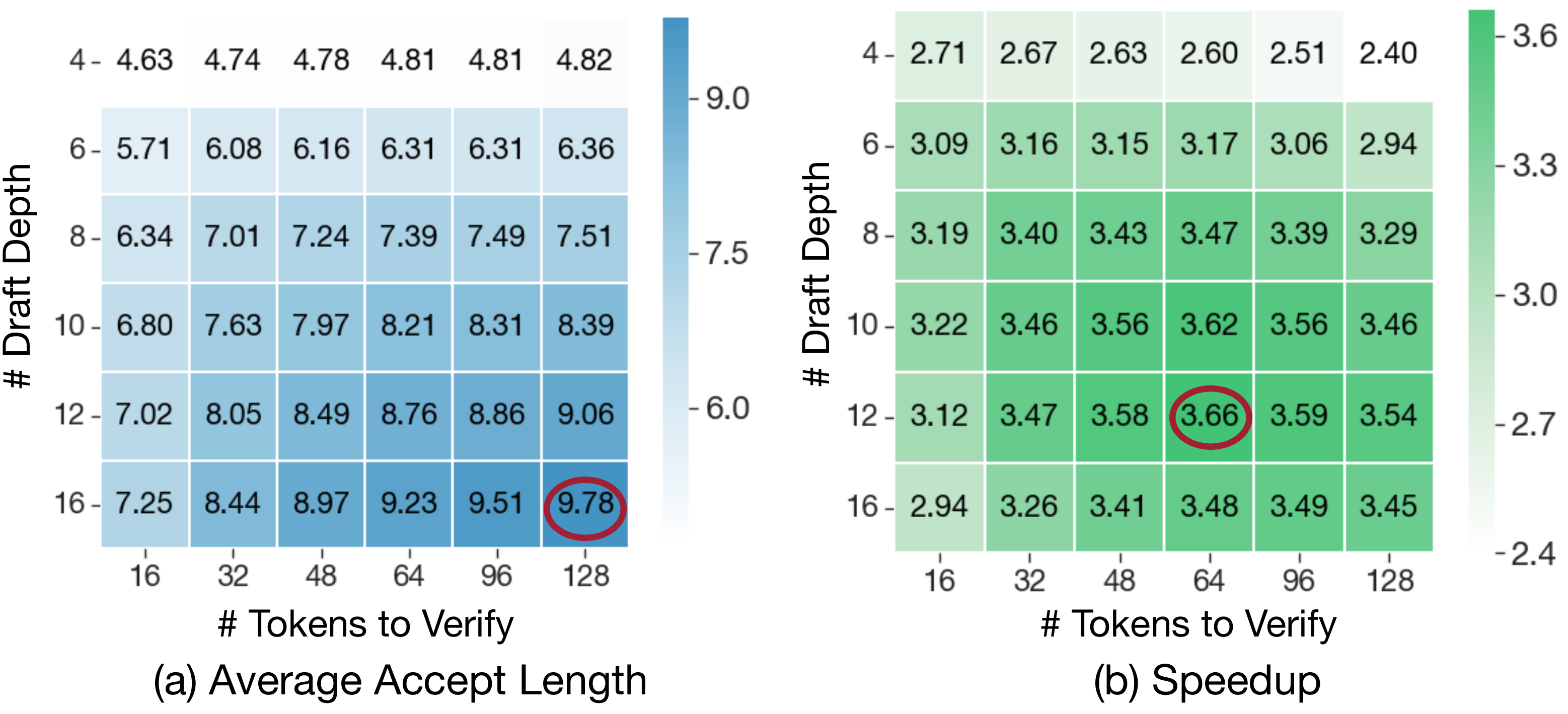}
    \caption{\textbf{Effect of hyperparameters for speculative decoding.} 
     Measurements performed with Qwen-32B (TP=4).}
    \label{fig:results:sd_heatmap}
\end{figure}

\noindent\textbf{Metrics}.
Following~\cite{HybridFlow}, we use token throughput (tokens/sec) for end-to-end evaluation. This metric is calculated by dividing the total number of tokens in both prompts and responses within a global batch by the duration of a single RL step. We average results over three training steps following an initial warm-up step and use a trained draft model for evaluation.

\subsection{End-to-End Evaluation}
\label{subsec_end2end}

We evaluate \SysName against popular RL training frameworks, including Open-R1~\cite{openr1} and  VeRL~\cite{HybridFlow}, across multiple GPU platforms (NVIDIA H100 and A100) and various LLM scales.  Figure~\ref{fig:results:main} demonstrates that \SysName consistently outperforms VeRL. Open-R1 consistently underperforms, primarily because it struggles to fully utilize all available GPUs concurrently and suffers from a tight coupling between its rollout and training batch size. Similar substantial gains are observed on the A100 platform, highlighting its effectiveness across different hardware generations. We also present results for \SysName-Base, which utilizes a model-free drafter and yields good benefits, demonstrating that the \SysName framework provides continuous efficiency gains throughout the entire training process. These end-to-end performance gains stem directly from \SysName's ability to mitigate the long-tail rollout bottleneck of reasoning RL training.

Moreover, Figure~\ref{fig:results:rl_trace} illustrates the average reward curves during RL training for both VeRL and \SysName using Qwen2.5-7B and Qwen2.5-32B models. It is evident that two curves overlap closely, indicating that \SysName preserves model quality. This confirms that \SysName's adaptive drafter and rollout engine effectively accelerate training without compromising the learning dynamics of the underlying RL algorithm.

\vspace{-10pt}
\subsection{Evaluation of Adaptive SD} 
\label{subsec_ablation}

\noindent\textbf{Effect of Adaptive SD Strategy}.
Tuning the SD configuration is crucial for achieving optimal speedups. Figure~\ref{fig:results:sd_heatmap} illustrates how \texttt{Draft\_Depth} and \texttt{Tokens\_to\_Verify} influence speculative decoding performance with batch size = 1, topK=8 and CUDA Graph enabled. For more stable accept length in the grid search, we set temperature=0. Increasing the draft depth generally raises the average accept length (a), though this benefit tapers off once the draft depth becomes sufficiently large. (b) reports the resulting speedups over the non-speculative baseline. This demonstrates that maximizing accept length alone is insufficient; drafting parameters must be tuned to optimize real performance rather than intermediate metrics.

Regarding \texttt{topK}, we find that efficiency is relatively insensitive to the value used during drafting (Table~\ref{tab:results:sd_topk}). To balance overall efficiency with framework simplicity, we fix this hyperparameter for our MAB auto-tuner.

\begin{table}[t]
\centering
\caption{\textbf{Effect of topK for speculative decoding.} Results are benchmarked using Qwen-32B on
H100 GPUs (TP=4), with \texttt{Draft\_Depth}=12 and \texttt{Tokens\_to\_Verify}=64.}

\scalebox{0.9}{
\begin{tabular}{ccccccc}

\toprule
TopK & 4  & 6  & 8    & 10 & 12    & 16   \\ 
\midrule
Accept Length & 8.29 & 8.66 & 8.67 & 8.67 & 8.60 & 8.42  \\ 
\midrule			
Speedup & 3.51$\times$ & \textbf{3.65}$\times$ & 3.64$\times$ & 3.64$\times$ & 3.56$\times$ & 3.47$\times$ \\ 
\bottomrule
\end{tabular}
}
\label{tab:results:sd_topk}
\end{table} 

Moreover, the optimal configurations for speculative decoding also vary with input batch size (Table~\ref{tab:results:sd_bs}). Although the speedup decreases as batch size grows, the model still benefits substantially from SD even at batch size 32. Another key observation is that \texttt{Tokens\_to\_Verify} should be adjusted based on the batch size: larger batches achieve optimal performance when fewer tokens are verified. These findings further demonstrate the effectiveness of our BEG MAB Selector in identifying appropriate configurations across dynamic running request numbers.

\begin{figure}[t]
    \centering
    \includegraphics[width=0.9\linewidth]{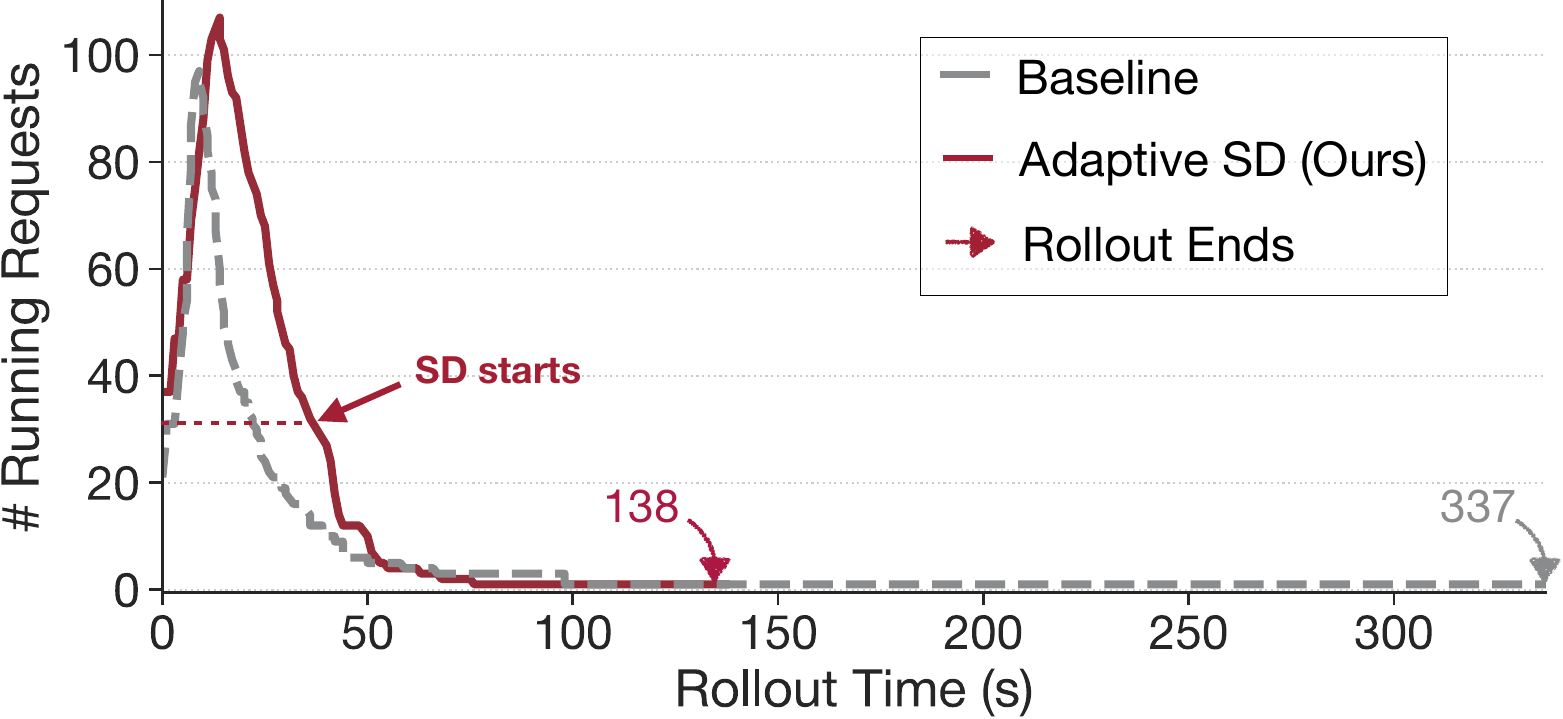}
    \caption{\textbf{Rollout running-request profiling.} Example of processing 128 requests with and without adaptive decoding using Qwen-32B (TP=4).}
    \label{fig:results:rollout_profile}
    \vspace{-15pt}
\end{figure}

\noindent\textbf{Case Study for Adaptive SD}.
Figure~\ref{fig:results:rollout_profile} presents a case study of the rollout profile during an RL step with 128 requests using Qwen-32B on H100 GPUs. Instead of consistently applying SD throughout the entire rollout process, leading to potential slowdown during the early phase when the number of running requests is high, \SysName switches to SD only when the number of remaining requests drops below a threshold (default 32), where SD becomes beneficial. Within these SD-enabled regions, \SysName further dynamically adapts SD configurations based on the current request count to maximize throughput. Overall, \SysName delivers a 2.44$\times$ speedup over the baseline system without speculative decoding.

\noindent\textbf{GPU Diversity and Scalability.} 
We evaluate \SysName across diverse GPU architectures and cluster scales. As in Table~\ref{tab:gpu_diversity}, \SysName delivers consistent speedups on both data-center and consumer GPUs. Furthermore, end-to-end training results (Table~\ref{tab:node_scaling}) demonstrate that \SysName's acceleration becomes more pronounced as model size and cluster size increase, reflecting favorable scalability for large-scale reasoning RL training.

\begin{table}[t]
\centering
\caption{\textbf{Rollout throughput (tokens/s) and speedup across GPU types }(Qwen2.5-7B, BS=1, TP=1).}
\label{tab:gpu_diversity}
\scalebox{0.9}{
\begin{tabular}{ccrc}
\toprule
\textbf{GPU Type} & \textbf{w/ SD} & \textbf{w/o SD} & \textbf{Speedup} \\
\midrule
B200     & 605.05 & 259.71 & 2.33$\times$ \\
H100     & 430.24 & 164.65 & 2.61$\times$ \\
A100     & 259.01 & 92.83  & 2.79$\times$ \\
RTX 5090 & 293.84 & 100.89 & 2.91$\times$ \\
RTX 4090 & 187.44 & 65.28  & 2.87$\times$ \\
RTX 3090 & 166.41 & 51.75  & 3.22$\times$ \\
\bottomrule
\end{tabular}
}
\vspace{-10pt}
\end{table}

\begin{table}[t]
\centering
\caption{\textbf{End-to-end training speed across GPU scales.}}
\label{tab:node_scaling}
\scalebox{0.8}{
\begin{tabular}{ccccc}
\toprule
\textbf{Nodes (8 GPUs/Node)} & \textbf{1 Node} & \textbf{2 Nodes} & \textbf{4 Nodes} & \textbf{8 Nodes} \\
\midrule
Qwen2.5-7B   & 1.21$\times$ & 1.45$\times$ & 1.62$\times$ & 1.76$\times$ \\
Qwen2.5-32B  & OOM          & OOM          & 1.83$\times$ & 2.12$\times$ \\
\bottomrule
\end{tabular}
}
\end{table}

\noindent\textbf{Bucketed CUDAGraph Capturing}. Preparing CUDAGraph for multiple SD strategies can be memory-intensive. As shown in Table~\ref{tab:results:cuda_graph_mem}, naively capturing a separate graph for each of the four candidate strategies inflates memory usage from 7.81 GB to 30.39 GB. Our Bucketed CUDAGraph design addresses this problem by reducing graph memory consumption to just 10.69 GB, a 2.8$\times$ reduction compared to the naive method and only a marginal increase over the single static strategy, allowing the adaptive tuner to flexibly switch between SD configurations for maximized efficiency.

\begin{table}[t]
\centering
\caption{\textbf{Effect of batch sizes on speculative decoding.} Speedup numbers are benchmarked using Qwen-32B on
H100 GPUs (TP=4), with \texttt{Draft\_Depth}=10 and \texttt{topK}=8.}

\scalebox{0.9}{
\begin{tabular}{cccccc}
\toprule
\multirow{2.5}{*}{\textbf{Speedup}} & \multicolumn{4}{c}{\textbf{\# Tokens to Verify}} \\
\cmidrule(lr){2-5}
 & \multicolumn{1}{c}{16} & \multicolumn{1}{c}{32} & \multicolumn{1}{c}{48} & \multicolumn{1}{c}{64}    \\
\midrule
Batch Size = 1       & 3.22$\times$  & 3.46$\times$  & 3.56$\times$           &  \textbf{3.62$\times$}  \\
Batch Size = 2       & 3.08$\times$  & 3.28$\times$  & \textbf{3.39$\times$}          &  3.38$\times$  \\
Batch Size = 4       & 3.01$\times$  & 3.09$\times$  & \textbf{3.13$\times$}         &  2.98$\times$  \\
Batch Size = 8       & \textbf{2.73$\times$}  & 2.63$\times$  & 2.51$\times$     & 2.27$\times$  \\
Batch Size = 16       & \textbf{2.67$\times$}  & 2.52$\times$  & 2.24$\times$     & 1.91$\times$  \\
Batch Size = 32       & \textbf{2.48$\times$}  & 2.23$\times$  & 1.90$\times$     & 1.70$\times$  \\
\bottomrule
\end{tabular}
}
\vspace{-10pt}
\label{tab:results:sd_bs}

\end{table} 

\vspace{-5pt}

\subsection{Evaluation of Spot Trainer}

\begin{figure}[t]
    \centering
    \includegraphics[width=0.9\linewidth]{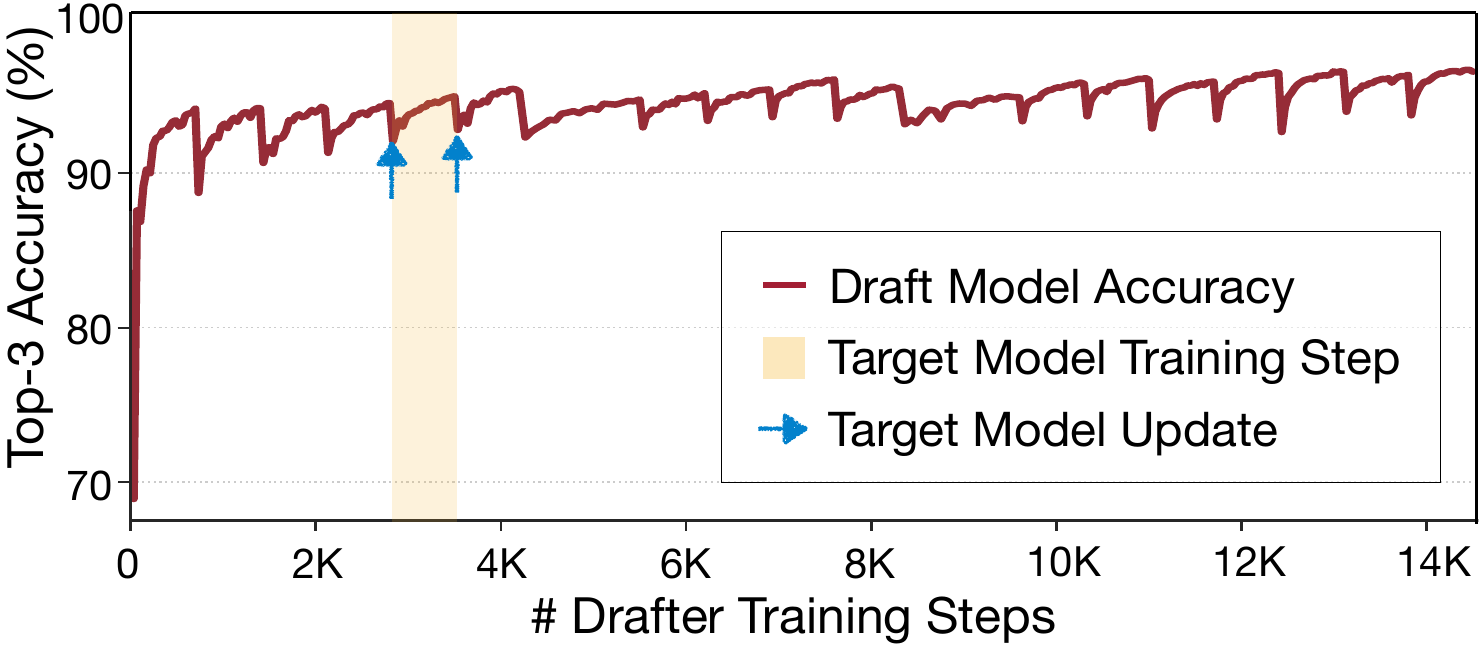}
    \caption{\textbf{Accuracy of the draft model during adaptive training.} The adaptive drafter's top-3 accuracy shows a consistent upward trend. While target model updates cause minor dips, the drafter quickly adapts and recovers. %
    }
    \vspace{-12pt}
    \label{fig:results:training_curve}
\end{figure}

\noindent\textbf{Training Adaptive Drafter}.
Figure~\ref{fig:results:training_curve} presents an example training curve illustrating the adaptive drafter's training process. After initial warm-up, the drafter's top-3 next-token prediction accuracy increases significantly. When the target model is updated after an RL step, the drafter's accuracy temporarily dips due to the distribution shift caused by the update. However, this accuracy is rapidly recovered within a few subsequent drafter training iterations, demonstrating the effectiveness and robustness of our adaptive method.

\noindent\textbf{Effectiveness of Adaptive Drafter}.
Table \ref{tab:results:eagle_trans} highlights the effectiveness of adaptively training the draft model. With this adaptive training mechanism, \SysName consistently achieves higher average accept lengths for the Target-R model across both RL-training tasks and downstream evaluations, demonstrating improved alignment and performance.

\begin{table}[t]

\centering
\caption{\textbf{Bucketed CUDAGraph reduces memory footprint.} Numbers measured using Llama-3-8B on H100 GPUs (TP=4). Search space includes 4 different strategies.}

\scalebox{0.85}{
\begin{tabular}{cc}
\toprule
\textbf{Method} & \textbf{Memory Footprint} \\
\midrule
Single Strategy & 7.81 GB \\
Vanilla Multiple Strategies & 30.39 GB \\
Bucketed CUDAGraph & 10.69 GB \\ 
\bottomrule
\end{tabular}
}
\label{tab:results:cuda_graph_mem}
\end{table} 

\begin{figure}[t]
    \centering
    \vspace{-10pt}
    \includegraphics[width=0.75\linewidth]{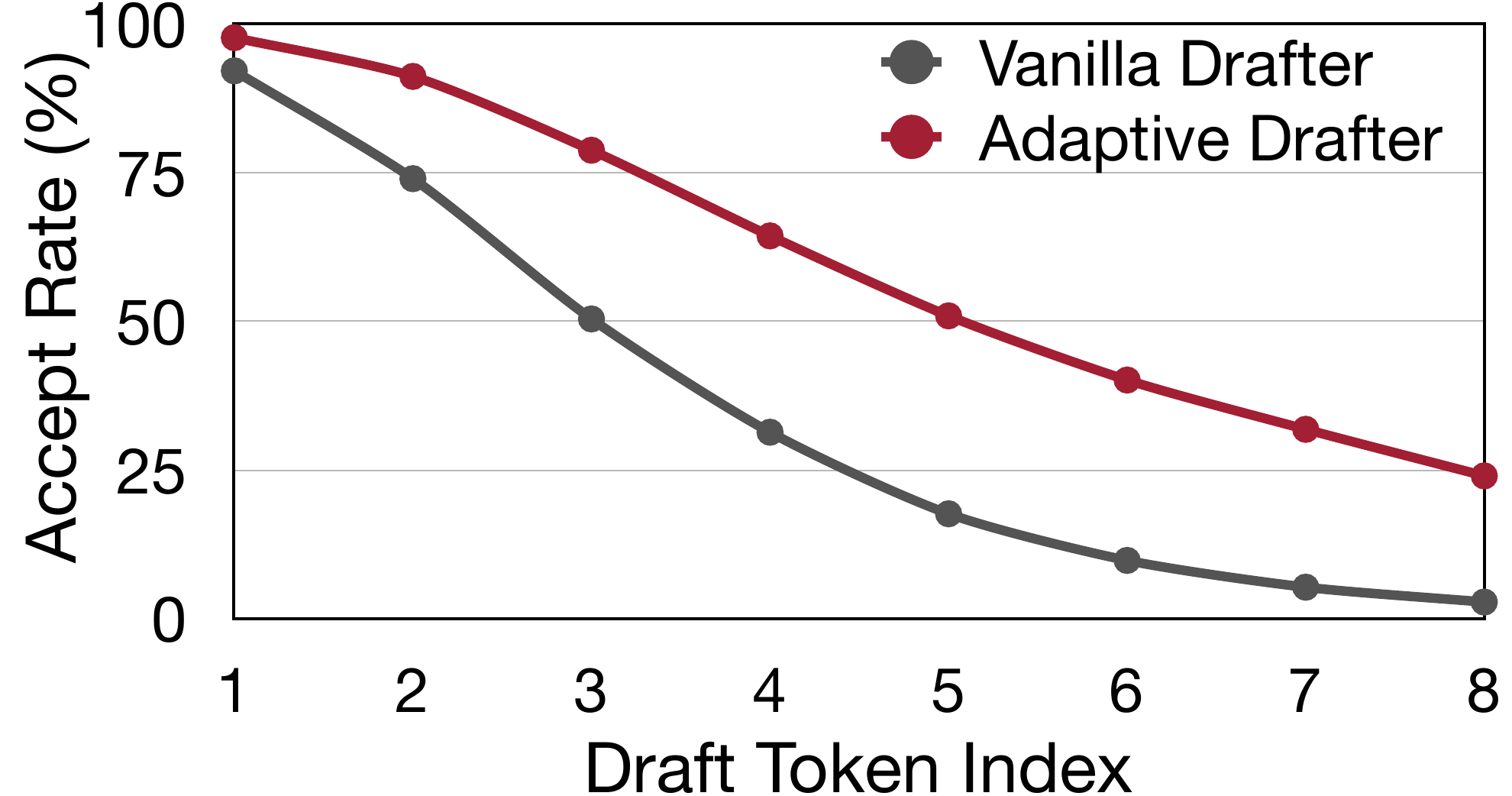}
    \caption{\textbf{Token accept rate of different draft models}. Our adaptive draft model achieves prominently higher accept rate when predicting distant tokens in rollout scenario.}
    \label{fig:results:alpha_ablate}
\end{figure}

Figure~\ref{fig:results:alpha_ablate} further corroborates this finding, which shows the acceptance probability at different token positions within the drafted sequence. Due to the output distribution shift of the target model and error accumulation, the vanilla draft model struggles to accurately predict tokens several steps ahead, severely limiting the effective accept length. The adaptive drafter, however, sustains higher acceptance probabilities across these positions, showcasing its ability to mitigate evolving target model and maintain longer accept lengths.

\begin{figure}[t]
    \centering
    \includegraphics[width=0.95\linewidth]{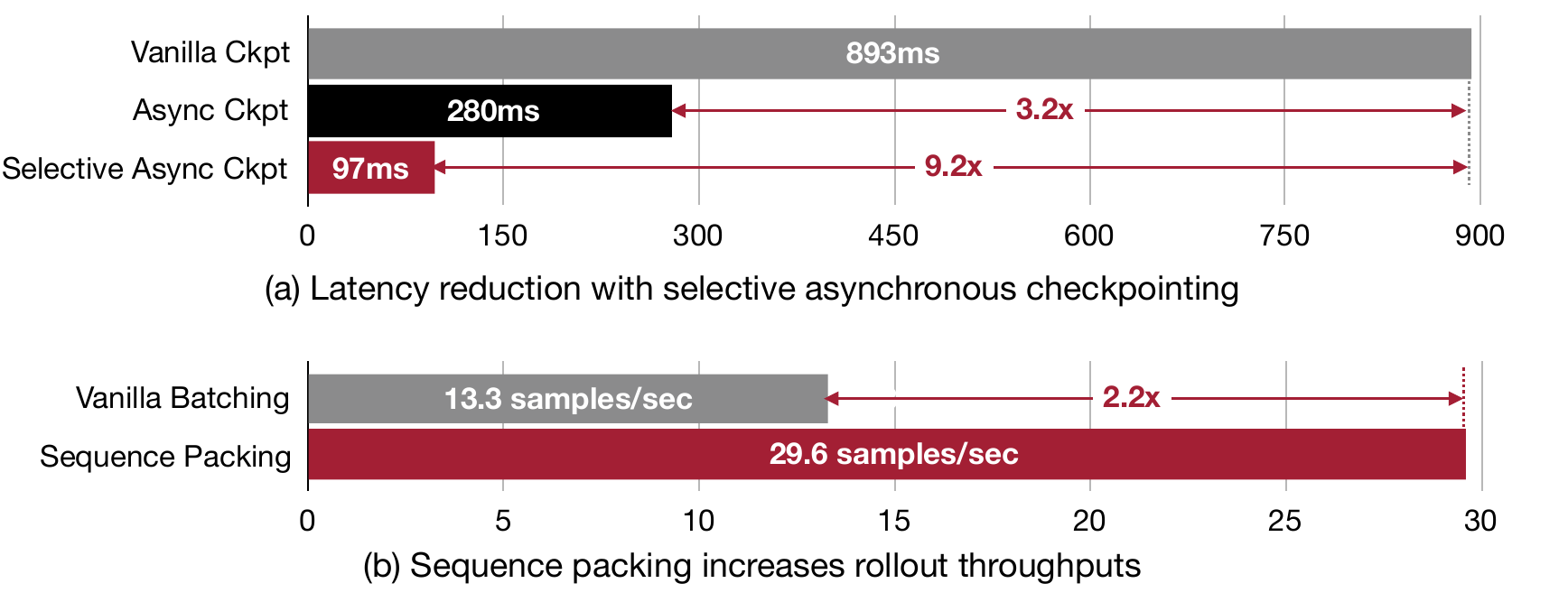}
    \caption{\textbf{Effect of selective asynchronous checkpointing and sequence packing.}}
    \label{fig:results:spot_trainer}
    \vspace{-5pt}
\end{figure}

\noindent\textbf{Efficient Spot Trainer}.
Spot Trainer achieves preemptible, high-throughput training with negligible runtime overheads. Among its optimizations, Selective Asynchronous Checkpointing (Figure~\ref{fig:results:spot_trainer} (a)) facilitates the trainer's preemptible design without significant work loss. Compared to the vanilla synchronous design, our approach reduces the checkpointing latency by 9.2 $\times$ through both offloading the save process to a background thread and selectively dumping only the trainable parameters of the draft model. Figure~\ref{fig:results:spot_trainer} (b) demonstrates the effectiveness of sequence packing. This technique eliminates wasteful computation on padding tokens by concatenating variable-length sequences, improving training throughput by 2.2$\times$ over vanilla batching.

\begin{table}[t]
\centering
\vspace{-5pt}
\caption{\textbf{Effectiveness of Adaptive Drafter.} 
Adaptive drafter maintains alignment with the target model. 
Target-Base denotes Qwen2.5-7B; and Target-R denotes the model after RL. Downstream indicates the collection of math, coding and reasoning domains in MTBench \cite{Arena}.
}
\scalebox{0.78}{
\begin{tabular}{ccccc}
\toprule

& \multicolumn{2}{c}{\textbf{RL Training}}      & \multicolumn{2}{c}{\textbf{Downstream}}                           \\
\cmidrule(lr){2-3}
\cmidrule(lr){4-5}
    & Target-Base & Target-R & Target-Base & Target-R \\ 
\midrule
Accept Length & 4.59                 & 6.53              & 3.76                 & 5.15    \\   
\bottomrule
\end{tabular}
}
\label{tab:results:eagle_trans}
\end{table} 

\label{sec_more_eval}

\noindent\textbf{System Overhead Analysis}.
We identify three primary sources of overhead in \SysName: (1) Stage-transition overhead, introduced by the additional drafter-weight update and optimizer offloading compared to VeRL \cite{HybridFlow}. This accounts for less than 1\% of the RL-step duration due to the much smaller model size. (2) SD switch overhead, incurred when switching from normal decoding to SD. Because the draft model is initially inactive for large batch sizes, we perform a reprefill step to activate SD, which completes within 3 seconds. (3) Drafter-training coordination overhead, caused by occasional contention during spot-trainer execution. However, we find that it is unnecessary to train the drafter every RL step, performing once every 10 steps is sufficient to maintain accuracy. Overall, the performance gains of \SysName far outweigh this overhead, yielding end-to-end speedups.

\vspace{-10pt}

\subsection{Analysis on Drafter Architectures} 
To demonstrate the generality of the \SysName framework, we integrate and evaluate several state-of-the-art speculative decoding methods in Table~\ref{tab:sd_ablation}. While newer SD methods like HASS~\cite{HASS} and Eagle-3~\cite{Eagle3} offer slightly higher accept length and throughput, their "Training-Time Test" technique, which requires multiple forward passes during the training process, significantly increases the training overhead per step and the difficulty of model convergence. Given the limited time budget in the rollout bubble, we chose Eagle~\cite{Eagle2} for its comparable performance and simplicity.

\begin{table}[t]
\centering
\caption{Comparison of different SD methods in TLT (Qwen2.5-7B, TP=2, BS=1, on H100 GPU). Training cost is normalized to per-step Eagle's cost.}
\label{tab:sd_ablation}
\scalebox{0.78}{
\begin{tabular}{lcccc}
\toprule
\textbf{Method} & \textbf{Accept Len} & \textbf{Throughput} & \textbf{Speedup} & \textbf{Training Cost} \\
\midrule
Base (No-SD)    & 1.00                & 242.11              & 1.00$\times$      & -                      \\
HASS~\cite{HASS}  & 6.67                & 553.91              & 2.29$\times$      & 3$\times$              \\
Eagle-3~\cite{Eagle3} & 6.83               & 617.42              & 2.55$\times$      & 7$\times$              \\
Eagle~\cite{Eagle2} & 6.53      & 542.12     & 2.24$\times$ & 1$\times$ \\
\bottomrule
\end{tabular}
}
\end{table}

\vspace{-10pt}
\subsection{Discuss Online SD (OSD)}
OSD~\cite{OSD} leverages general knowledge distillation (GKD) for the drafter model so it remains aligned with distribution shifts in online serving. In contrast, \SysName focuses on RL training, and the target model's weights are continuously evolving. While KD is already incorporated as a loss term in Eagle training, OSD further explores reverse KD. We investigate the impact of different training paradigms in Table~\ref{tab:osd_impact}. The results demonstrate that OSD-style training more effectively improves draft model quality than supervised fine-tuning (SFT) for a general small model (e.g., Qwen-0.5B), as well as the default Eagle training method used in TLT.

\begin{table}[t]
\vspace{-12pt}
\centering
\small
\caption{Impact of OSD-style training on different draft models (Qwen2.5-7B target, H100 GPU, $TP=2, BS=1$).}
\label{tab:osd_impact}
\resizebox{\linewidth}{!}{
\begin{tabular}{@{}ccccccc@{}}
\toprule
\textbf{\begin{tabular}[c]{@{}c@{}}Draft\\ Model\end{tabular}} & \textbf{\begin{tabular}[c]{@{}c@{}}Original\\ Accpet Len\end{tabular}} & \textbf{\begin{tabular}[c]{@{}c@{}}Original\\ Thpt.\end{tabular}} & \textbf{\begin{tabular}[c]{@{}c@{}}Trained \\ Accpet Len\end{tabular}} & \textbf{\begin{tabular}[c]{@{}c@{}}Trained \\ Thpt.\end{tabular}} & \textbf{\begin{tabular}[c]{@{}c@{}}+OSD\\ Accpet Len\end{tabular}} & \textbf{\begin{tabular}[c]{@{}c@{}}+OSD\\ Thpt.\end{tabular}} \\ \midrule
Qwen2.5-0.5B                                                   & 5.95                                                                   & 363.87                                                                 & 6.68                                                                      & 408.51                                                                      & 6.89                                                             & 421.35                                                           \\
Eagle                                                          & 1.57                                                                   & 130.34                                                                 & 6.53                                                                   & 542.12                                                                 & 6.77                                                                & 562.04                                                                \\ \bottomrule
\end{tabular}}
\end{table}

\vspace{-5pt}
\section{Discussion}
\label{sec_discussion}

\noindent\textbf{More Application Scenarios}. 
In this work, we primarily investigate how speculative decoding can accelerate RL training by mitigating long-tail rollouts. We believe this technique also opens promising avenues for broader applications. For example:
(1) \textit{Uniformly long responses}. When all rollouts are long and no tail exists, each request demands substantial KV-cache capacity. This often makes the system KV-cache-bound and triggers request eviction, limiting the running batch size. In such cases, the workload naturally falls into the “sweet spot" of speculative decoding, enabling decoding acceleration.
(2) \textit{Multi-turn rollouts with tool-calling RL}. In multi-turn RL settings involving tool calls, partial requests perform GPU-free tool executions while their KV caches remain resident on the GPU. This reduces the number of active decoding requests and again creates a favorable regime for speculative decoding under our approach.
(3) \textit{Online serving and edge deployment}. When the target model is fixed, a draft model trained via \SysName can be directly deployed for inference. Adaptive speculative decoding remains highly effective in these deployment scenarios, especially for handling variable load. Extending our approach to these settings is an exciting direction for future work.

\noindent\textbf{Can We Break RL Synchronization}?
\SysName aims to improve RL training efficiency without modifying on-policy requirements of the underlying algorithm.
One may wonder whether it is possible to further accelerate training by asynchronously updating the model with partial rollouts during long decoding phases. While appealing, this approach risks altering the underlying RL algorithm and potentially harming convergence and model performance. Asynchronous RL with or continuous batching~\cite{AReaL} typically mixes multiple policy versions within a single rollout, potentially biasing gradient estimation and shifting the learning dynamics (e.g., drifting from on-policy to off-policy). Additional algorithmic modifications are required to safely accommodate such paradigm. A more balanced approach is to allow limited asynchrony that preserves algorithmic correctness while improving hardware utilization. Even under such settings, the heavy-tail rollout problem remains, and our speculative decoding approach can be combined with limited asynchronous RL to further accelerate training \emph{without} introducing additional loss or policy staleness.

  
%

\noindent\textbf{Can \SysName be applied to other RL algorithms}?
GRPO represents a widely used algorithmic structure for reasoning RL. Alternatives such as RLOO~\cite{RLOO}, DAPO~\cite{DAPO}, REINFORCE~\cite{REINFORCE}, and REINFORCE++~\cite{REINFORCEpp} generally share the same overall training workflow, differing mainly in their reward formulations and use of KL-regularization. These algorithmic variations remain fully compatible with our adaptive drafter and spot-training design, suggesting that \SysName can readily accelerate a broad class of RL methods.

\vspace{-9pt}
\section{Related Work}
\label{sec_related}
\noindent\textbf{RL Systems for LLMs}. Several systems aim to optimize the training pipeline of Reinforcement Learning from Human Feedback (RLHF) \cite{InstructGPT}. Frameworks like \cite{DeepSpeedChat, GEAR, HybridFlow, FlexRLHF, NeMoAligner} provide scalable end-to-end solutions, supporting various alignment algorithms and parallelism techniques. RLHFuse~\cite{RLHFuse} employs fine-grained scheduling and stage fusion to improve GPU utilization. ReaLHF~\cite{ReaLHF} introduces dynamic parameter reallocation to flexibly adapt 3D parallelism strategies across RLHF stages. Their optimization efforts primarily target the efficient management of multiple models inherent in the RLHF process. However, they neglect a critical performance bottleneck in the rollout phase. This oversight renders existing systems inefficient for reasoning RL tasks. Our work addresses this gap by concentrating specifically on optimizing the rollout bottleneck, utilizing an adaptive drafter that is orthogonal to prior systems. Recent efforts \cite{AReaL, StreamRL, zhou2025april} attempt to alleviate the synchronization constraint in RL by allowing stale responses for model updates or terminating rollout prematurely. RollPacker \cite{gao2025rollpacker} introduces tail batching; however, even for identical queries, reasoning trajectories can vary substantially in length, limiting its effectiveness. While these strategies accelerate training, they risk degrading model quality.  In contrast, our approach preserves the original RL algorithm, ensuring lossless training performance while achieving significant efficiency gains.


\noindent\textbf{Speculative Decoding}.
Speculative decoding~\cite{SpeculativeDecoding, chen2023accelerating, AdaServe} accelerates LLM inference by verifying candidate tokens in parallel with the target model, effectively increasing the number of tokens generated per step. Enhancements to speculative decoding encompass diverse strategies. Researchers have explored retrieval-based methods~\cite{REST, saxena2023PLD, lookahead}, which avoid the need to train an extra draft model. Others modify the target model directly: The Medusa series~\cite{Medusa} integrates multiple prediction heads, an approach refined by Hydra~\cite{Hydra} which models correlations between these predictions. Furthermore, SpecInfer~\cite{SpecInfer} innovatively employs tree-based speculative decoding to broaden the candidate search space. OSD \cite{OSD} leverages knowledge distillation for the drafter model so it remains aligned with distribution shifts in online serving.  Additionally, EAGLE~\cite{Eagle1, Eagle2, Eagle3} trains separate, compact auto-regressive models as dedicated draft models. Further refining this approach, HASS~\cite{HASS} improves draft model accuracy by simulating multi-step predictions during training, mitigating the training-inference inconsistency issue. Recent work also adapts speculative ideas for efficient reasoning LLM inference. For example, RSD~\cite{RSD} employs a reward model to guide the speculation, and SpecReason~\cite{SpecReason} uses a lightweight model to speculatively execute intermediate reasoning steps. Unlike these approaches designed for inference with static models, we focus on the dynamic training scenario where the target model is continuously updated.

\vspace{-15pt}
\section{Conclusion}
\label{sec_conclusion}

To conclude, \SysName alleviates the long-tail bottleneck in reasoning RL training via an adaptive drafter, delivering substantial throughput gains without compromising model quality.



\section*{Acknowledgements}
We sincerely thank our shepherd, Prof. Jiarong Xing, for his guidance throughout the paper revision process. We also thank the anonymous reviewers of ASPLOS and SOSP for their insightful and valuable feedback.
We thank MIT-IBM Watson AI Lab, MIT AI Hardware Program, MIT Amazon Science Hub, Hyundai Motor Company, and National Science Foundation for supporting this work.

\bibliographystyle{ACM-Reference-Format}%
\bibliography{references.bib}

\end{document}